\documentclass{article} 

\PassOptionsToPackage{numbers, compress}{natbib}

\usepackage[preprint]{neurips_2024}

\usepackage[utf8]{inputenc} 
\usepackage[T1]{fontenc}    
\usepackage{url}            
\usepackage{booktabs}       
\usepackage{amsfonts}       
\usepackage{nicefrac}       
\usepackage{microtype}      
\usepackage{xcolor}         
\usepackage{graphicx}
\usepackage{subfigure}
\usepackage{amsmath}
\usepackage{multirow}
\usepackage{multicol}
\usepackage{array}
\usepackage{makecell}

\usepackage{amssymb}
\usepackage{booktabs}
\usepackage{multirow}
\usepackage{bm}
\usepackage{bbding}

\usepackage{wrapfig} 
\usepackage{xcolor} 

\definecolor{citecolor}{HTML}{0071bc}
\usepackage[colorlinks, linkcolor=red, colorlinks, anchorcolor=blue, citecolor=citecolor, pagebackref=True]{hyperref}

\usepackage{authblk}

\usepackage{soul}

\usepackage{colortbl}
\usepackage[capitalize]{cleveref}
\crefname{section}{Sec.}{Secs.}
\Crefname{section}{Section}{Sections}
\Crefname{table}{Table}{Tables}
\crefname{table}{Tab.}{Tabs.}

\newcommand{\ourMethod}{LION}

\title{LION: Linear Group RNN for 3D Object\\ Detection in Point Clouds}

\author[ ]{\textbf{Zhe Liu}$^{1*}$}
\author[ ]{\textbf{Jinghua Hou}$^{1*}$}
\author[ ]{\textbf{Xinyu Wang}$^{1}$\thanks{Equal contribution.}}
\author[ ]{\textbf{Xiaoqing Ye}$^{3}$}
\author[ ]{\textbf{Jingdong Wang}$^{3}$}
\author[ ]{\authorcr \textbf{Hengshuang Zhao}$^{2}$}
\author[ ]{\textbf{Xiang Bai}$^{1}$\thanks{Corresponding author.}}
{\affil[1]{Huazhong University of Science and Technology} \affil[2]{The University of Hong Kong \ \ }}
\affil[3]{Baidu Inc., China}
\affil[ ]{\textcolor{magenta}{\url{https://happinesslz.github.io/projects/LION}}}

\begin{document}

\maketitle

\begin{abstract}


The benefit of transformers in large-scale 3D point cloud perception tasks, such as 3D object detection, is limited by their quadratic computation cost when modeling long-range relationships. In contrast, linear RNNs have low computational complexity and are suitable for long-range modeling.
Toward this goal, we propose a simple and effective window-based framework built on \textbf{LI}near gr\textbf{O}up RN\textbf{N}~(\textit{i.e.}, perform linear RNN for grouped features) for accurate 3D object detection, called \textbf{{\ourMethod}}. The key property is to allow sufficient feature interaction in a much larger group than transformer-based methods. However, effectively applying linear group RNN to 3D object detection in highly sparse point clouds is not trivial due to its limitation in handling spatial modeling. To tackle this problem, we simply introduce a 3D spatial feature descriptor and integrate it into the linear group RNN operators to enhance their spatial features rather than blindly increasing the number of scanning orders for voxel features. 
To further address the challenge in highly sparse point clouds, we propose a 3D voxel generation strategy to densify foreground features thanks to linear group RNN as a natural property of auto-regressive models. 
Extensive experiments verify the effectiveness of the proposed components and the generalization of our {\ourMethod} on different linear group RNN operators including Mamba, RWKV, and RetNet. Furthermore, it is worth mentioning that our {\ourMethod}-Mamba achieves state-of-the-art on Waymo, nuScenes, Argoverse V2, and ONCE datasets. Last but not least, our method supports kinds of advanced linear RNN operators~(\textit{e.g.}, RetNet, RWKV, Mamba, xLSTM and TTT) on small but popular KITTI dataset for a quick experience with our linear RNN-based framework.




\end{abstract}

\section{Introduction}


\begin{figure*}[ht]
\centering
\includegraphics[width=0.99\linewidth]{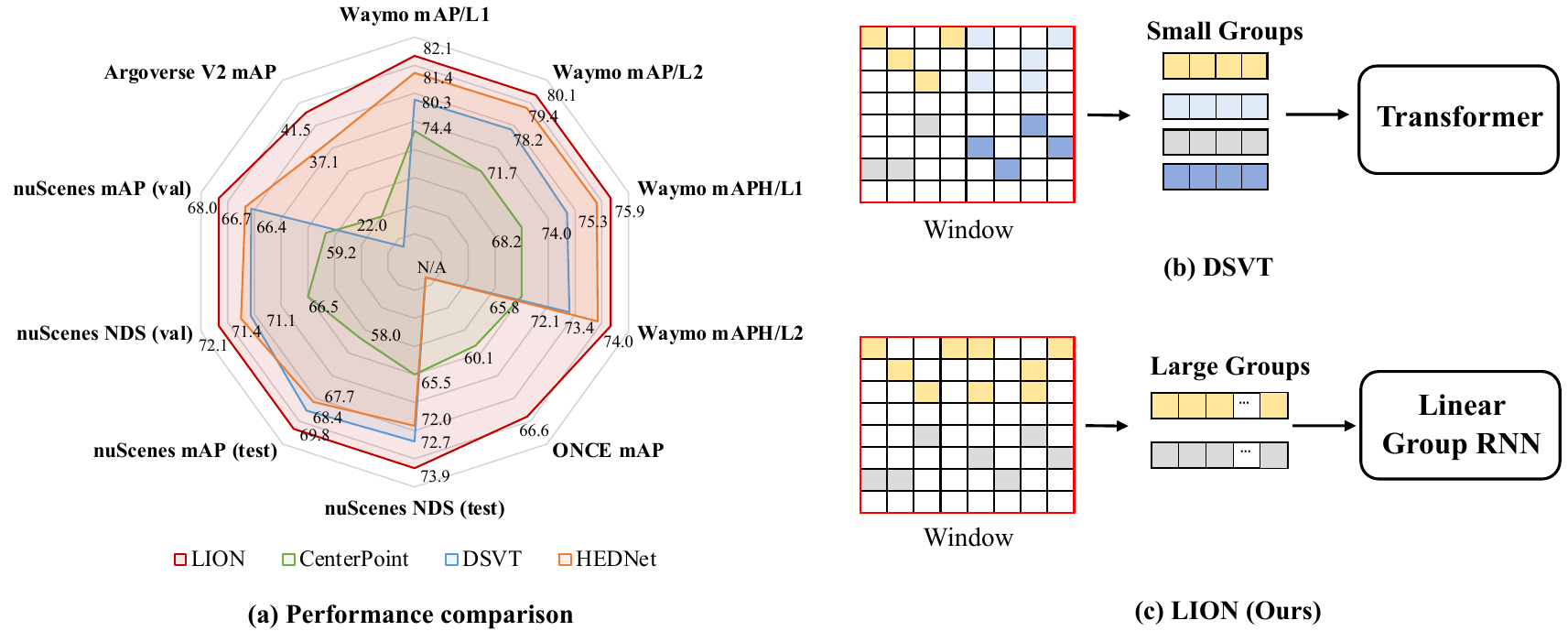}
\vspace{-5pt}
\caption{(a)~Comparison of different 3D backbones in terms of detection performance on Waymo~\cite{sun2020scalability}, nuScenes~\cite{caesar2020nuscenes}, Argoverse V2~\cite{wilson2023argoverse} and ONCE~\cite{mao2021one} datasets. Here, we adopt Mamba~\cite{gu2023mamba} as the default operator of our {\ourMethod}. Besides, we present the simplified schematic of DSVT~(b)~\cite{wang2023dsvt} and our {\ourMethod}~(c) for implementing feature interaction in 3D backbones.}
\label{fig_intro}
\vspace{-15pt}
\end{figure*}

3D object detection serves as a fundamental technique in 3D perception and is widely used in navigation robots and self-driving cars.
Recently, transformer-based~\cite{vaswani2017attention} feature extractors have made significant progress in general tasks of Natural Language Processing (NLP) and 2D vision by flexibly modeling long-range relationships. To this end, some researchers have made great efforts to transfer the success of transformers to 3D object detection. Specifically, to reduce the computation costs, SST~\cite{fan2022embracing} and SWFormer~\cite{Sun2022SWFormerSW} divide point clouds into pillars and implement window attention for pillar feature interaction in a local 2D window. Considering some potential information loss of the pillar-based manners along the height dimension, DSVT-Voxel~\cite{wang2023dsvt} further adopts voxel-based formats and implements set attention for voxel feature interaction in a limited group size.

Although the above methods have achieved some success in 3D detection, they perform self-attention for pillar or voxel feature interaction with only a small group size due to computational limitations, locking the potential of transformers for modeling long-range relationships. Moreover, it is worth noting that modeling long-range relationships can benefit from large datasets, which will be important for achieving foundational models in 3D perception tasks in the future. 
Fortunately, in the field of large language models~(LLM) and 2D perception tasks, some representative linear RNN operators such as  Mamba~\cite{gu2023mamba} and RWKV~\cite{peng2023rwkv} with linear computational complexity have achieved competitive performance with transformers, especially for long sequences.
Therefore, a question naturally arises: can we perform long-range feature interaction in larger groups at a lower computation cost based on linear RNNs in 3D object detection? 

To this end, we propose a window-based framework based on \textbf{LI}near gr\textbf{O}up RN\textbf{N}~(\textit{i.e.}, perform linear RNN for grouped features in a window-based framework) termed \textbf{\ourMethod} for accurate 3D object detection in point clouds. Different from the existing method DSVT~(b) in Figure~\ref{fig_intro}, our {\ourMethod}~(c) could support thousands of voxel features to interact with each other in a large group for establishing the long-range relationship. 
Nevertheless, effectively adopting linear group RNN to construct a proper 3D detector in highly sparse point cloud scenes remains challenging for capturing the spatial information of objects.
Concretely, linear group RNN requires sequential features as inputs. However, converting voxel features into sequential features may result in the loss of spatial information~(\textit{e.g.}, two features that are close in 3D spatial position might be very far in this 1D sequence).
Therefore, we propose a simple 3D spatial feature descriptor and decorate the linear group RNN operators with it, thus compensating for the limitations of linear group RNN in 3D local spatial modeling.

Furthermore, to enhance feature representation in highly sparse point clouds, we present a new 3D voxel generation strategy based on linear group RNN to densify foreground features. 
A common manner of addressing this is to add an extra branch to distinguish the foregrounds, as seen in previous methods~\cite{Sun2022SWFormerSW,fan2022fully,zhang2024safdnet}. However, this solution is relatively complex and rarely used in 3D backbone due to its lack of structural elegance. Instead, we simply choose the high response of the feature map in the 3D backbone as the areas for voxel generation. Subsequently, the auto-regressive property of linear group RNN can be effectively employed to generate voxel features.

Finally, as shown in Figure~\ref{fig_intro} (a), we compare {\ourMethod} with the existing representative methods. We can clearly observe that our {\ourMethod} achieves state-of-the-art on a board autonomous datasets in terms of detection performance. To summarize, our contributions are as follows:
\textbf{1)} We propose a simple and effective  window-based 3D backbone based on the linear group RNN named {\ourMethod} to allow long-range feature interaction.
\textbf{2)} We introduce a simple 3D spatial feature descriptor and integrate it with the linear group RNN, compensating for the lack of capturing 3D local spatial information.
\textbf{3)} We provide a new 3D voxel generation strategy to densify foreground features, producing a more discriminative feature representation in highly sparse point clouds.
\textbf{4)} We verify the generalization of our {\ourMethod} with different linear group RNN mechanisms~(\textit{e.g.}, Mamba, RWKV, RetNet). In particular, our {\ourMethod}-Mamba achieves state-of-the-art on challenging Waymo~\cite{sun2020scalability}, nuScenes~\cite{caesar2020nuscenes}, Argoverse V2~\cite{wilson2023argoverse}, and ONCE~\cite{mao2021one} dataset, which further illustrates the superiority of {\ourMethod}.


\section{Related Work}

\noindent \textbf{3D Object Detection in Point Clouds.}~3D object detectors in point clouds can be roughly divided into point-based and voxel-based. For point-based methods~\cite{chen2019fast, yang2019std, qi2019deep,Cheng_2021_CVPR,liu2021group,pan20213d,he2020structure, shi2019pointrcnn,zhang2022not,yang20203dssd,qi2018frustum,yang2022dbq,chen2022sasa}, they usually sample point clouds and adopt point encoder~\cite{qi2017pointnet,qi2017pointnet++} to directly extract point features. However, the point sampling and grouping utilized by point-based methods is time-consuming. To avoid these problems, voxel-based methods~\cite{dong2022mssvt,deng2021voxel,liu2020tanet, shi2020pv,shi2021pv,shi2020points,guan2022m3detr,wang2022cagroup3d, yan2018second,yin2021center,yang2023pvt,zhang2024safdnet} convert the input irregular point clouds into regular 3D voxels and then extract 3D features by 3D sparse convolution. Although these methods achieve promising performance, they are still limited by the local receptive field of 3D convolution. Therefore, some methods~\cite{chen2023largekernel3d,lu2023link} adopt the large kernel to enlarge the receptive field and achieve better performance. 

\noindent \textbf{Linear RNN.}~Recurrent Neural Networks (RNNs) are initially developed to address problems in Natural Language Processing (NLP), such as time series prediction and speech recognition, by effectively capturing temporal dependencies in sequential data. 
Recently, to overcome the quadratic computational complexity of transformers, significant advancements have been made in time-parallelizable data-dependent RNNs~(called linear RNNs in this paper)~\cite{qin2023hierarchically,orvieto2023resurrecting,peng2023rwkv, peng2024eagle, sun2023retentive, de2024griffin, yang2023gated, gu2023mamba,sun2024learning,beck2024xlstm}. These models retain linear complexity while offering efficient parallel training capabilities, allowing their performance to match or even surpass that of transformers. 
Due to their scalability and efficiency, linear RNNs are poised to play an increasingly important role in various fields and some works~\cite{duan2024vision, alkin2024vision,liang2024pointmamba} have applied linear RNNs to 2D/3D vision filed. 
In this paper, we aim to further extend linear RNNs to 3D object detection tasks thanks to their long-range relationship modeling capabilities.


\noindent \textbf{Transformers in 3D Object Detection.}~Transformer~\cite{vaswani2017attention} has achieved great success in many tasks, motivating numerous works to adopt attention mechanisms in 3D object detection to achieve better performance. However, the application of transformers is non-trivial in large-scale point clouds.
Many works \cite{Fan_2022_CVPR, Sun2022SWFormerSW, liu2023flatformer, wang2023dsvt} apply transformers to extract features by partitioning pillars or voxels into several groups based on local windows. Although these approaches achieve promising performance,  they usually adopt small groups for feature interaction due to the quadratic computational complexity of transformers, hindering them from capturing long-range dependencies in 3D space. In contrast, we propose a simple and effective framework based on linear RNNs named {\ourMethod} to achieve long-range feature interaction for accurate 3D object detection thanks to their linear computational complexity.


\section{Method}

\begin{figure*}[t]
\centering
\includegraphics[width=0.99\linewidth]{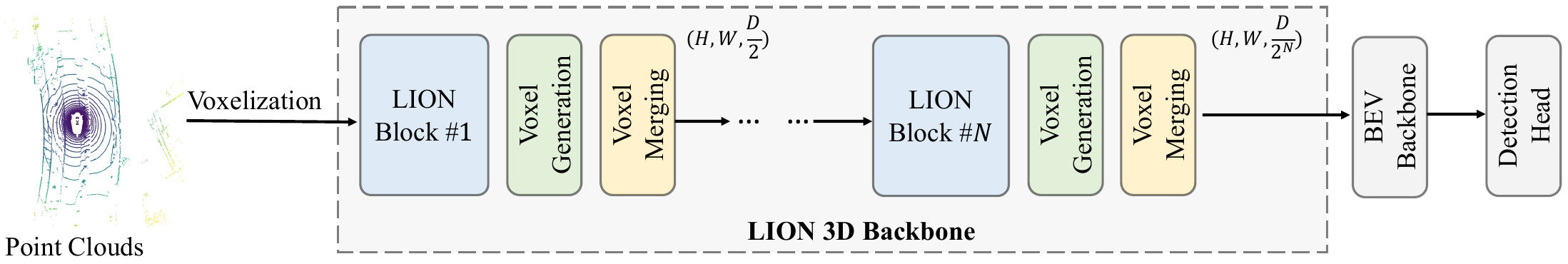}
\caption{The illustration of {\ourMethod}, which mainly consists of several {\ourMethod} blocks, each paired with a voxel generation for feature enhancement and a voxel merging for down-sampling features along the height dimension. $(H, W, D)$ indicates the shape of the 3D feature map, where $H$, $W$, and $D$ are the length, width, and height of the 3D feature map along the X-axis, Y-axis, and Z-axis. $N$ is the number of {\ourMethod} blocks. In {\ourMethod}, we first convert point clouds to voxels and partition these voxels into a series of equal-size groups. Then, we feed these grouped features into {\ourMethod} 3D backbone to enhance their feature representation. Finally, these enhanced features are fed into a BEV backbone and a detection head for final 3D detection. 
}
\label{fig_pipeline}
\vspace{-5pt}
\end{figure*}


Due to computational limitations, some transformer-based methods~\cite{fan2022embracing,wang2023dsvt,liu2023flatformer} usually convert features into pillars or group small size of voxel features to interact with each other within small groups, limiting the advantages of transformers in long-range modeling. 
More recently, some linear RNN operators~\cite{gu2023mamba,peng2023rwkv,sun2023retentive} that maintain linear complexity with the length of the input sequence are proposed to model long-range feature interaction. More importantly, the linear RNN operators such as Mamba~\cite{gu2023mamba} and RWKV~\cite{peng2023rwkv} have even shown comparable performance with transformers in LLM thanks to their low computation cost in long-range feature interaction. This further motivates us to adopt linear RNNs to construct a 3D detector for long-range modeling, which might be meaningful for the unified multi-modal large model in 3D perception in the near future. 
However, effectively applying linear group RNN to 3D object detection is challenging and rarely explored due to the lack of strong spatial modeling in highly sparse point cloud scenes. Next, we will introduce our solution.


\subsection{Overview}

In this paper, we propose a simple and effective window-based framework based on \textbf{LI}near gr\textbf{O}up RN\textbf{N}~(\textit{i.e.}, perform linear RNN for grouped features in a window-based framework) named \textbf{{\ourMethod}}, which can group thousands of voxels (dozens of times more than the number of previous methods~\cite{fan2022embracing,wang2023dsvt,liu2023flatformer}) for feature interaction.
The pipeline of our {\ourMethod} is presented in Figure~\ref{fig_pipeline}. {\ourMethod} consists of a 3D backbone, a BEV backbone, and a detection head, maintaining a consistent pipeline with most voxel-based 3D detectors~\cite{yan2018second, wang2023dsvt, yin2021center}. In this paper, our contribution lies in the design of the 3D backbone based on linear group RNN. In the following, we will present the details of our proposed 3D backbone, which includes $N$ {\ourMethod} blocks for long-range feature interaction, $N$ voxel generation operations for enhancing feature representation in sparse point clouds, and $N$ voxel merging operations for gradually down-sampling features in height.

\noindent\textbf{3D Sparse Window Partition.} Our {\ourMethod} is a window-based 3D detector. Thus, before feeding voxel features into our {\ourMethod} block, we need to implement a 3D sparse window partition to group them for feature interaction. Specifically, 
we first convert point clouds into voxels with the total number of $L$.
Then, we divide these voxels into non-overlapping 3D windows with the shape of $(T_x, T_y, T_z)$, where $T_x$, $T_y$ and $T_z$ denote the length, width, and height of the window along the X-axis, Y-axis, and Z-axis. Next, we sort voxels along the X-axis for the X-axis window partition and along the Y-axis for the Y-axis window partition, respectively.
Finally, to save computation cost, we adopt the equal-size grouping manner in FlatFormer~\cite{liu2023flatformer} instead of the classic equal-window grouping manner in SST~\cite{fan2022embracing}. That is, we partition sorted voxels into groups 
with equal size $K$ rather than windows of equal shapes for feature interaction.
Due to the quadratic computational complexity of transformers, previous transformer-based methods~\cite{fan2022embracing,wang2023dsvt,liu2023flatformer} only achieve feature interaction using a small group size. In contrast, we adopt a much larger group size $K$ to obtain long-range feature interaction thanks to the linear computational complexity of the linear group RNN operators.

\begin{figure*}[t!]
\centering
\includegraphics[width=0.95\linewidth]{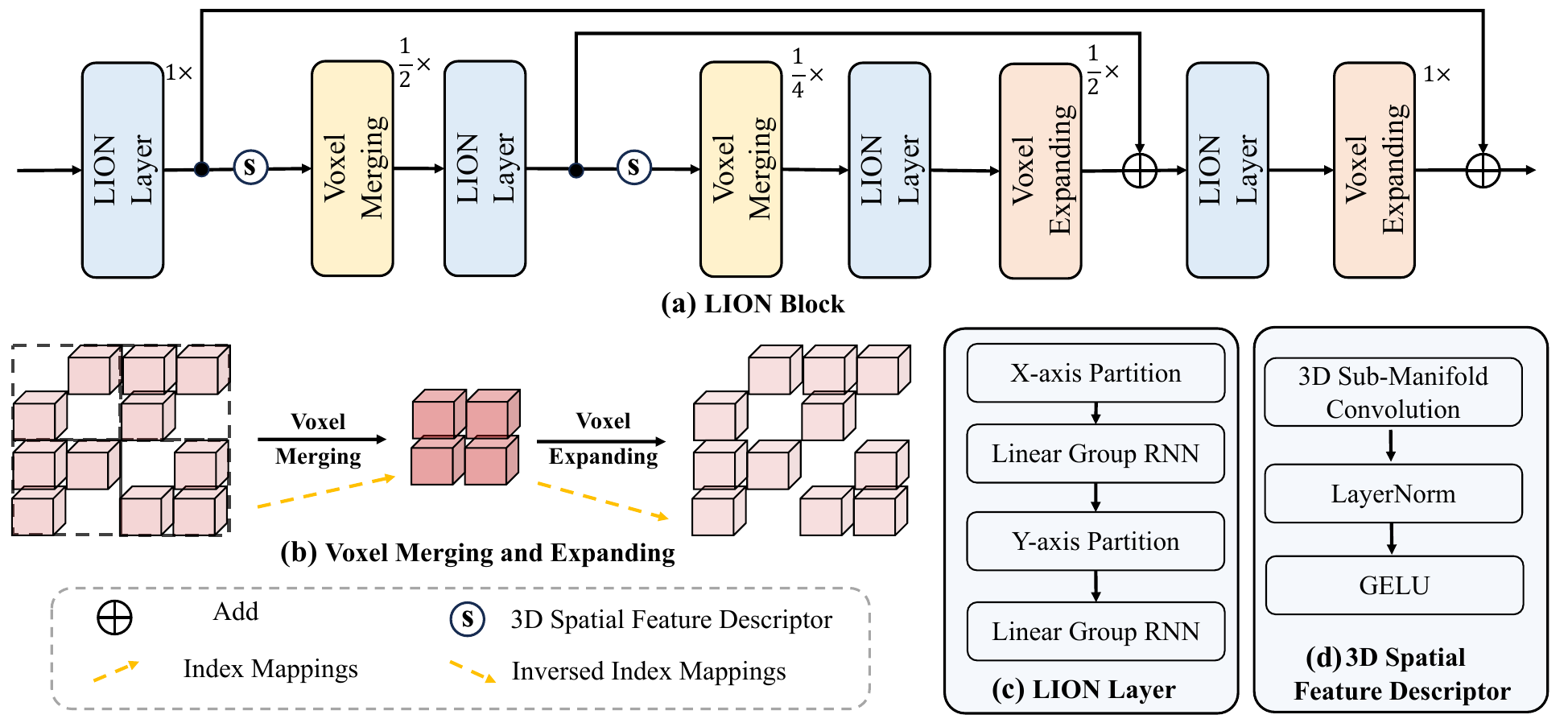}
\vspace{-5pt}
\caption{(a) shows the structure of {\ourMethod} block, which involves four {\ourMethod} layers, two voxel merging operations, two voxel expanding operations, and two 3D spatial feature descriptors. Here, $1\times$, $\frac{1}{2}\times$, and $\frac{1}{4}\times$ indicate the resolution of 3D feature map as $(H,W,D)$, $(H/2,W/2,D/2)$ and $(H/4,W/4,D/4)$, respectively.
(b) is the process of voxel merging for voxel down-sampling and voxel expanding for voxel up-sampling. (c) presents the structure of {\ourMethod} layer. (d) shows the details of the 3D spatial feature descriptor.  
}
\label{fig_block}
\vspace{-10pt}
\end{figure*}

\subsection{{\ourMethod} Block}
The {\ourMethod} block is the core component of our approach, which involves {\ourMethod} layer for long-range feature interaction, 3D spatial feature descriptor for capturing local 3D spatial information, voxel merging for
feature down-sampling and voxel expanding for feature up-sampling, as shown in Figure~\ref{fig_block}~(a). Besides, {\ourMethod} block is a hierarchical structure to better extract multi-scale features due to the gap of different 3D objects in size. Next, we introduce each part of {\ourMethod} block.

\noindent\textbf{{\ourMethod} Layer.} In {\ourMethod} block, we apply {\ourMethod} layer to model a long-range relationship among grouped features with the help of the linear group RNN operator. Specifically, as shown in Figure~\ref{fig_block}~(c), we provide the structure of {\ourMethod} layer, which is composed of two linear group RNN operators. The first one is used to perform long-range feature interaction based on the X-axis window partition and the second one can extract long-range feature information based on the Y-axis window partition.
Taking advantage of two different window partitions, {\ourMethod} layer can obtain more sufficient feature interaction, producing more discriminative feature representation.


\begin{wrapfigure}{r}{0.3\textwidth}
	\centering
  \vspace{-10pt}
	\includegraphics[width=\linewidth]{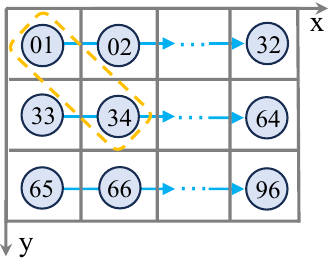}
	\caption{\small{
 The illustration of spatial information loss when flattening into 1D sequences. For example, there are two adjacent voxels in spatial position~(indexed as 01 and 34)  but are far in the 1D sequences along the X order.}
		}\label{fig:1d}
  \vspace{-5pt}
\end{wrapfigure}

\noindent\textbf{3D Spatial Feature Descriptor.} Although linear RNNs have the advantages of long-range modeling with low computation cost, it is not ignorable that the spatial information might be lost when input voxel features are flattened into 1D sequential features.
For example, as shown in Figure~\ref{fig:1d}, there are two adjacent features~(\textit{i.e.}, indexed as 01 and 34) in 3D space. However, after they are flattened into 1D sequential features, the distance between them in 1D space is very far. We regard this phenomenon as a loss of 3D spatial information. 
To tackle this problem, an available manner is to increase the number of scan orders for voxel features such as VMamba~\cite{liu2024vmamba}. However, the order of scanning is too hand-designed. Besides, as the scanning orders increase, the corresponding computation cost also increases significantly. Therefore, it is not appropriate in large-scale sparse 3D point clouds to adopt this manner.
As shown in Figure~\ref{fig_block}~(d), we introduce a 3D spatial feature descriptor, which consists of a 3D sub-manifold convolution, a LayerNorm layer, and a GELU activation function. Naturally, we can leverage the 3D spatial feature descriptor to provide rich 3D local position-aware information for the {\ourMethod} layer. Besides, we place the 3D spatial feature descriptor before the voxel merging to reduce spatial information loss in the process of voxel merging. We provide the corresponding experiment in our appendix.

\noindent\textbf{Voxel Merging and Voxel Expanding}. 
To enable the network to obtain multi-scale features, our {\ourMethod} adopts a hierarchical feature extraction structure. To achieve this, we need to perform feature down-sampling and up-sampling operations in highly sparse point clouds. However, it is worth mentioning that we cannot simply apply max or average pooling or up-sampling operations as in 2D images since 3D point clouds possess irregular data formats. 
Therefore, as shown in Figure~\ref{fig_block}~(b), we adopt voxel merging for feature down-sampling and voxel expanding for feature up-sampling in highly sparse point clouds. Specifically, for voxel merging, we calculate the down-sampled index mappings to merge voxels. In voxel expanding, we up-sample the down-sampled voxels by the corresponding inversed index mappings.




\subsection{Voxel Generation}
\begin{figure*}[h!]
\centering
\includegraphics[width=0.99\linewidth]{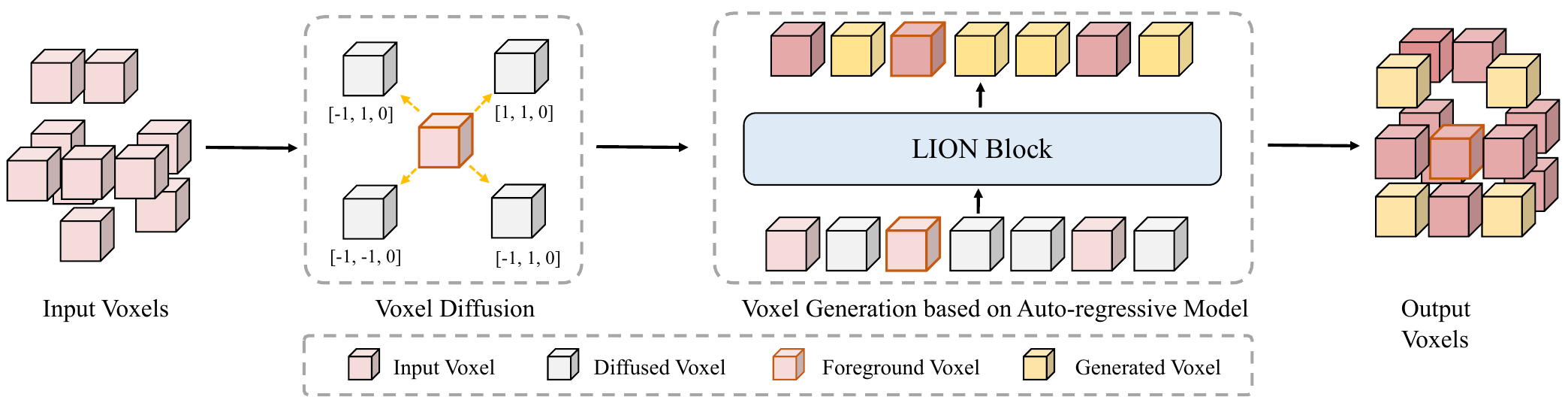}
\vspace{-5pt}
\caption{
The details of voxel generation. For input voxels, we first select the foreground voxels and diffuse them along different directions. Then, we initialize the corresponding features of the diffused voxels as zeros and utilize the auto-regressive ability of the following {\ourMethod} block to generate diffused features. Note that we do not present the voxel merging here for simplicity.
}
\label{fig_vfg}
\end{figure*}

Considering the challenge of feature representation in highly sparse point clouds and the potential information loss of implementing voxel merging in Figure~\ref{fig_pipeline}, we propose a voxel generation strategy to address these issues with the help of the auto-regressive capacity of the linear group RNN. 

\noindent\textbf{Distinguishing Foreground Voxels without Supervision.} 
 In voxel generation, the first challenge is identifying which regions of voxel features need to be generated. Different from previous methods~\cite{Sun2022SWFormerSW, fan2022fully, zhang2024safdnet} that employ some supervised information based on well-learned BEV features to obtain the foreground region for feature diffusion. However, these approaches may be unsuitable for a 3D backbone and may even compromise its elegance. 
Interestingly, we notice that the corresponding high values of feature responses along the channel dimension in the 3D backbone~(Refer to the visualization in Figure~\ref{fig_heatmap} in our appendix) are usually the foregrounds. Therefore, we compute the feature response $F_i^*$ for the output feature $F_i$ of the ${i}^{th}$ {\ourMethod} block, where $i=1,2,..., N$ indicates the index of {\ourMethod} block in the 3D backbone. Thus, this above process can be formulated as:
\begin{equation}
F_i^* = \frac{1}{C} \sum_{j=0}^{C}F_i^j,
\end{equation}
where $C$ is the channel dimension of $F_i$. Next, we sort the feature responses  $F_i^*$ in descending order and select the corresponding Top-$m$ voxels as the foregrounds from the total number $L$ of non-empty voxels, where $m = r *L$ and $r$ is the ratio of foregrounds. This process can be computed as:
\begin{equation}
{F}_{m} = \mathrm{Top}_{m}({F_i^*}),
\end{equation}
where $\mathrm{Top}_{m}({F_i^*})$ means selecting Top-$m$ voxel features from $F_i^*$. ${F}_{m}$ are the selected foreground features, which will serve for the subsequent voxel generation.

\noindent\textbf{Voxel Generation with Auto-regressive Property.} 
The previous method~\cite{Sun2022SWFormerSW} adopts a K-NN manner to obtain generated voxel features based on their K-NN features, which might be sub-optimal to enhance feature representation due to the redundant features and the limited receptive field. Fortunately, the linear RNN is well-suited for auto-regressive tasks in addition to its advantage of handling long sequences. Therefore, we leverage the auto-regressive property of linear RNN to effectively generate the new voxel features by performing sufficient feature interaction with other voxel features in a large group. 
Specifically, for convenience, we define the corresponding coordinates of selected foreground voxel features ${F}_{m}$ as $P_m$. As shown in Figure~\ref{fig_vfg}, 
we first obtain diffused voxels by diffusing $P_m$ with four different offsets (\textit{i.e.}, [-1,-1, 0], [1,1, 0], [1,-1, 0], and [-1,1, 0]) along the X-axis, Y-axis, and Z-axis, respectively. Then, we initialize the corresponding features of diffused voxels by all zeros. Next, we concatenate the output feature $F_i$ of the ${i}^{th}$ {\ourMethod} block with the initialized voxel features, and feed them into the subsequent $({i+1})^{th}$ {\ourMethod} block. Finally, thanks to the auto-regressive ability of the {\ourMethod} block, the diffused voxel features can be effectively generated based on other voxel features in large groups. This process can be formulated as:
\begin{equation}
{F}_{p} = F_i \oplus {F}_{[-1, -1, 0]} \oplus F_{[1, 1, 0]} \oplus F_{[1, -1, 0]} \oplus F_{[-1, 1, 0]}, 
\end{equation}
\begin{equation}
F^{'}_{p} = \mathrm{Block}(F_{p}), 
\end{equation}
where $F_{[x, y, z]}$ denotes the initialized voxel features with diffused offsets of x, y, and z along the X-axis, Y-axis, and Z-axis. The $\oplus$ and $\mathrm{Block}$ denote the concatenation and {\ourMethod} block respectively.

\section{Experiments}

\subsection{Datasets and Evaluation Metrics}
\noindent\textbf{Waymo Open Dataset.} 
Waymo Open dataset~(WOD)~\cite{sun2020scalability} is a well-known benchmark for large-scale outdoor 3D perception, comprising 1150 scenes which are divided into 798 scenes for training, 202 scenes for validation, and 150 scenes for testing. Each scene includes about 200 frames, covering a perception range of $150m \times 150m$. For evaluation metrics, WOD employs 3D mean Average Precision~(mAP) and mAP weighted by heading accuracy~(mAPH), each divided into two difficulty levels: L1 is for objects detected with more than five points and L2 is for those at least one point. 

\noindent\textbf{nuScenes Dataset.}  
nuScenes~\cite{caesar2020nuscenes} is a popular outdoor 3D perception benchmark with a perception range of up to 50 meters. Each frame in the scene is annotated with 2Hz.
The dataset includes 1000 scenes, which is divided into 750 scenes for training, 150 scenes for validation, and 150 scenes for testing. nuScenes adopts mean Average Precision~(mAP) and the NuScenes Detection Score~(NDS) as evaluation metrics. 

\noindent\textbf{Argoverse V2 Dataset.}
Argoverse V2~\cite{wilson2023argoverse} is a popular outdoor 3D perception benchmark with a long-range perception of up to 200 meters. It contains 1000 sequences in total, 700 for training, 150 for validation, and 150 for testing. Each frame in the scene is annotated with 10Hz. For the evaluation metric, Argoverse v2 adopts a similar mean Average Precision~(mAP) metric with nuScenes~\cite{caesar2020nuscenes}. 

\noindent\textbf{ONCE Dataset.}
ONCE~\cite{mao2021one} is another representative autonomous driving dataset, which consists of 5000, 3000, and 8000 frames for training, validation, and testing set, respectively. Each frame is annotated with 5 classes (Car, Bus, Truck, Pedestrian, and Cyclist). Besides, ONCE merges the car, bus, and truck class into a super-class called vehicle following WOD~\cite{sun2020scalability}. For the detection metric, ONCE extends~\cite{geiger2013vision} by taking the object orientations into special consideration and evaluating the final performance by mAP for three classes.

\subsection{Implementation Details}
\noindent\textbf{Network Architecture.} 
In our {\ourMethod}, we provide three representative linear RNN operators~(\textit{i.e.}, Mamba~\cite{gu2023mamba}, RWKV~\cite{peng2023rwkv}, and RetNet~\cite{sun2023retentive}). Each of operator adopts a bi-directional structure to better 
capture 3D geometric information inspired by ~\cite{graves2013speech}.
On WOD, we keep the same channel dimension $C=64$ for all {\ourMethod} blocks in {\ourMethod}-Mamba, {\ourMethod}-RWKV, and {\ourMethod}-RetNet. For the large version of {\ourMethod}-Mamba-L, we set $C=128$.
We follow DSVT-Voxel~\cite{wang2023dsvt} to set the grid size as (0.32m, 0.32m, 0.1875m). The number of {\ourMethod} blocks $N$ is set to 4. For these four {\ourMethod} blocks, the window sizes $(T_x, T_y, T_z)$ are set to $(13, 13, 32)$, $(13, 13, 16)$, $(13, 13, 8)$, and $(13, 13, 4)$, and the corresponding group sizes $K$ are {$4096$, $2048$, $1024$, $512$}, respectively. Besides, we adopt the same center-based detection head and loss function as DSVT~\cite{wang2023dsvt} for fair comparison. In the voxel generation, we set the ratio $r=0.2$ to balance the performance and computation cost. For the nuScenes dataset, we replace DSVT~\cite{wang2023dsvt} 3D backbone with our {\ourMethod} 3D backbone except for changing the grid size to $(0.32m, 0.32m, 0.15m)$. For the Argoverse V2 dataset, we replace the 3D backbones of VoxelNext~\cite{chen2023voxelnext} or SAFDNet~\cite{zhang2024safdnet} with our {\ourMethod} 3D backbone except for setting the grid size to $(0.4m, 0.4m, 0.25m)$. Moreover, it is noted that we only add  three extra LION layers to further enhance the 3D backbone features, rather than applying the BEV backbone to obtain the dense BEV features. For the ONCE dataset, we replace SECOND~\cite{yan2018second} 3D backbone with our {\ourMethod} 3D backbone except for adopting the grid size as $(0.4m, 0.4m, 0.25m)$. 


\noindent\textbf{Training Process.} 
On the WOD, we adopt the same point cloud range, data augmentations, learning rate, and optimizer as the previous method~\cite{wang2023dsvt}. We train our model 24 epochs with a batch size of 16 on 8 NVIDIA Tesla V100 GPUs.
Besides, we utilize the fade strategy~\cite{wang2021pointaugmenting} to achieve better performance in the last epoch. For the nuScenes dataset, we adopt the same point cloud range, data augmentations, and optimizer as previous method~\cite{bai2022transfusion}. Moreover, we find that {\ourMethod} converges faster than previous methods on nuScenes dataset. Therefore, we only train our model for 36 epochs without CBGS~\cite{zhu2019class}. The learning rate and batch size are set to 0.003 and 16, respectively. It is worth noting that the CBGS strategy extends training iterations about 4.5 times, which means that our training iterations are much fewer than previous methods~\cite{bai2022transfusion} (\textit{i.e.}, 20 epochs with CBGS). For the Argoverse V2 dataset and the ONCE dataset, we adopt the same training process with SAFDNet~\cite{zhang2024safdnet} and SECOND~\cite{yan2018second}, respectively.

\subsection{Main Results}
In this section, we provide a board comparison of our {\ourMethod} with existing methods on  WOD, nuScenes, Argoverse V2 and ONCE datasets for 3D object detection. Furthermore, in the section ~\ref{sec:kitti} of our appendix, we present more types of linear RNN operators~(\textit{e.g.}, RetNet, RWKV, Mamba, xLSTM, and TTT) based on our {\ourMethod} framework for 3D detection on a small but popular dataset KITTI~\cite{geiger2012we} for a quick experience.

\noindent\textbf{Results on WOD.} 
To illustrate the superiority of our {\ourMethod}, we provide the comparison with 
existing representative methods on the WOD in Table~\ref{tab:waymo_main_results}. Here, we also conduct the experiments on our {\ourMethod} with different linear group RNN operators, including {\ourMethod}-Mamba, {\ourMethod}-RWKV and {\ourMethod}-RetNet. Compared with the transformer-based methods~\cite{fan2022embracing,Sun2022SWFormerSW,wang2023dsvt}, our {\ourMethod} with different linear group RNN operators outperforms the previous state-of-the-art~(SOTA) transformer-based 3D backbone DSVT-Voxel~\cite{wang2023dsvt}, illustrating the generalization of our proposed framework. To further scale up our {\ourMethod}, we present the performance of {\ourMethod}-Mamba-L by doubling the channel dimension of {\ourMethod}-Mamba. It can be observed that {\ourMethod}-Mamba-L significantly outperforms DSVT-Voxel with 1.9 mAPH/L2, leading to a new SOTA performance. The above promising results effectively demonstrate the superiority of our proposed {\ourMethod}.




\begin{table}[t!]
\caption{Performances on the Waymo Open Dataset \textit{validation} set (train with 100\% training data). $\ddag$ denotes the two-stage method. \textbf{Bold} denotes the best performance of all methods. ``-L" means we double the dimension of channels in {\ourMethod} 3D backbone. RNN denotes the linear RNN operator. All results are presented with single-frame input, no test-time augmentation, and no model ensembling.
}
\vspace{-5pt}
\footnotesize
\centering
\resizebox{1.0\linewidth}{!}{
\begin{tabular}{l|c|c|c|c|c|c|c|c|>{\columncolor[gray]{0.95}}c}
\specialrule{1pt}{1pt}{0pt}
\multicolumn{1}{c|}{\multirow{2}{*}{Methods}} & \multirow{2}{*}{Present at} & \multirow{2}{*}{Operator}  & \multicolumn{2}{c|}{\emph{Vehicle} 3D AP/APH} & \multicolumn{2}{c|}{\emph{Pedestrian} 3D AP/APH} & \multicolumn{2}{c|}{\emph{Cyclist} 3D AP/APH} & \multirow{2}{*}{\shortstack[1]{mAP/mAPH \\ L1}} \\
 & & & L1 & L2 & L1 & L2 & L1 & L2 & L2 \\
\midrule
SECOND~\cite{yan2018second} & Sensors 18 & \multirow{15}{*}{\rotatebox{90}{Sparse Conv}}  & 72.3/71.7 & 63.9/63.3 & 68.7/58.2 & 60.7/51.3 & 60.6/59.3 & 58.3/57.0 & 61.0/57.2 \\
PointPillars~\cite{lang2019pointpillars} & CVPR 19 &  & 72.1/71.5 & 63.6/63.1 & 70.6/56.7 & 62.8/50.3 & 64.4/62.3 & 61.9/59.9 & 62.8/57.8\\ 
CenterPoint~\cite{yin2021center}& CVPR 21 &  & 74.2/73.6 & 66.2/65.7 & 76.6/70.5 & 68.8/63.2 & 72.3/71.1 & 69.7/68.5 & 68.2/65.8 \\
PV-RCNN$\ddag$~\cite{shi2020pv} & CVPR 20 &  & 78.0/77.5 & 69.4/69.0 & 79.2/73.0 & 70.4/64.7 & 71.5/70.3 & 69.0/67.8 & 69.6/67.2\\
PillarNet-34~\cite{shi2022pillarnet} & ECCV 22 &  & 79.1/78.6 & 70.9/70.5 & 80.6/74.0 & 72.3/66.2 & 72.3/71.2 & 69.7/68.7 & 71.0/68.5 \\
FSD$\ddag$~\cite{fan2022fully} & NeurIPS 22 &  & 79.2/78.8 & 70.5/70.1 & 82.6/77.3 & 73.9/69.1 & 77.1/76.0 & 74.4/73.3 & 72.9/70.8 \\
AFDetV2~\cite{hu2022afdetv2} & AAAI 22 &  & 77.6/77.1 & 69.7/69.2 & 80.2/74.6 & 72.2/67.0 & 73.7/72.7 & 71.0/70.1 & 71.0/68.8 \\
PillarNeXt~\cite{li2023pillarnext} & CVPR 23 &  & 78.4/77.9 & 70.3/69.8 & 82.5/77.1 & 74.9/69.8 & 73.2/72.2 & 70.6/69.6 & 71.9/69.7\\ 
VoxelNext~\cite{chen2023voxelnext} & CVPR 23 &  & 78.2/77.7 & 69.9/69.4 & 81.5/76.3 & 73.5/68.6 & 76.1/74.9 & 73.3/72.2 & 72.2/70.1 \\ 
CenterFormer\cite{zhou2022centerformer} & ECCV 22 &  & 75.0/74.4 & 69.9/69.4 & 78.6/73.0 & 73.6/68.3 & 72.3/71.3 & 69.8/68.8 & 71.1/68.9\\ 
PV-RCNN++$\ddag$~\cite{shi2021pv} & IJCV 22 &  & 79.3/78.8 & 70.6/70.2 & 81.3/76.3 & 73.2/68.0 & 73.7/72.7 & 71.2/70.2 & 71.7/69.5\\ 
TransFusion~\cite{bai2022transfusion} & CVPR 22 &  & \textbf{--}/\textbf{--} & \textbf{--}/65.1 & \textbf{--}/\textbf{--} & \textbf{--}/63.7 & \textbf{--}/\textbf{--} & \textbf{--}/65.9 & \textbf{--}/64.9 \\ 
ConQueR~\cite{zhu2023conquer} & CVPR 23 &  & 76.1/75.6 & 68.7/68.2 & 79.0/72.3 & 70.9/64.7 & 73.9/72.5 & 71.4/70.1 & 70.3/67.7 \\
FocalFormer3D~\cite{chen2023focalformer3d} & ICCV 23 &  & \textbf{--}/\textbf{--} & 68.1/67.6 & \textbf{--}/\textbf{--} & 72.7/66.8 & \textbf{--}/\textbf{--} & 73.7/72.6 & 71.5/69.0\\ 
HEDNet~\cite{zhang2023hednet} & NeurIPS 23 &  & 81.1/80.6 & 73.2/72.7 & 84.4/80.0 & 76.8/72.6 & 78.7/77.7 & 75.8/74.9 & 75.3/73.4 \\
\midrule
SST\_TS$\ddag$~\cite{fan2022embracing} & CVPR 22 &\multirow{5}{*}{\rotatebox{90}{Transformer}} & 76.2/75.8 & 68.0/67.6 & 81.4/74.0 & 72.8/65.9 & \textbf{--}/\textbf{--} & \textbf{--}/\textbf{--}  & \textbf{--}/\textbf{--} \\
SWFormer~\cite{Sun2022SWFormerSW} & ECCV 22 &  & 77.8/77.3 & 69.2/68.8 & 80.9/72.7 & 72.5/64.9 & \textbf{--}/\textbf{--} & \textbf{--}/\textbf{--} & \textbf{--}/\textbf{--} \\
OcTr~\cite{zhou2023octr} & CVPR 23 &  & 78.1/77.6 & 69.8/69.3 & 80.8/74.4 & 72.5/66.5 & 72.6/71.5 & 69.9/68.9 & 70.7/68.2\\ 
DSVT-Pillar~\cite{wang2023dsvt} & CVPR 23 &  & 79.3/78.8 & 70.9/70.5 & 82.8/77.0 & 75.2/69.8 & 76.4/75.4 & 73.6/72.7 & 73.2/71.0\\ 
DSVT-Voxel~\cite{wang2023dsvt} & CVPR 23 & & 79.7/79.3 & 71.4/71.0 & 83.7/78.9 & 76.1/71.5 & 77.5/76.5 & 74.6/73.7 & 74.0/72.1\\ 
\midrule

{\ourMethod}-RetNet~(Ours) & \textbf{--} &\multirow{4}{*}{\rotatebox{90}{RNN}} & 79.0/78.5 & 70.6/70.2 & 84.6/80.0 & 77.2/72.8 & 79.0/78.0 & 76.1/75.1 & 74.6/72.7 \\ 
{\ourMethod}-RWKV~(Ours) & \textbf{--} & & 79.7/79.3 & 71.3/71.0 & 84.6/80.0 & 77.1/72.7 & 78.7/77.7 & 75.8/74.8 & 74.7/72.8 \\ 
{\ourMethod}-Mamba~(Ours) & \textbf{--} & & 79.5/79.1 & 71.1/70.7 & 84.9/80.4 & 77.5/73.2 & 79.7/78.7 & 76.7/75.8 & 75.1/73.2 \\ 
{\ourMethod}-Mamba-L~(Ours) & \textbf{--} & & \textbf{80.3/79.9} & \textbf{72.0/71.6} & \textbf{85.8/81.4} & \textbf{78.5/74.3} & \textbf{80.1/79.0} & \textbf{77.2/76.2} & \textbf{75.9/74.0} \\ 
\bottomrule
\end{tabular}
}
\vspace{-10pt}
\label{tab:waymo_main_results}
\end{table}

\begin{table}[t]
\caption{Performances on the nuScenes \textit{validation} and \textit{test} set. ‘T.L.’, ‘C.V.’, ‘Ped.’, ‘M.T.’, ‘T.C.’, and 'B.R.' are short for trailer, construction vehicle, pedestrian, motor, traffic cone, and barrier, respectively. All results are reported without any test-time augmentation and model ensembling.
}
\vspace{-5pt}
\small
\setlength{\tabcolsep}{6pt}
\resizebox{1.0\linewidth}{!}{
\centering
\begin{tabular}{l|c|cc|cccccccccc}
\toprule
\multicolumn{14}{c}{Performances on the \textit{validation} set}\\
\midrule
Method & Present at & NDS & mAP & Car & Truck & Bus & T.L. & C.V. & Ped. & M.T. & Bike & T.C. & B.R.\\
\midrule
CenterPoint~\cite{yin2021center} & CVPR 21 & 66.5 & 59.2 & 84.9 & 57.4 & 70.7 & 38.1 & 16.9 & 85.1 & 59.0 & 42.0 & 69.8 & 68.3 \\
VoxelNeXt~\cite{chen2023voxelnext} & CVPR 23 & 66.7 & 60.5 & 83.9 & 55.5 & 70.5 & 38.1 & 21.1 & 84.6 & 62.8 & 50.0 & 69.4 & 69.4 \\
Uni3DETR~\cite{wang2024uni3detr}  & NeurIPS 23 & 68.5 & 61.7 & \textbf{--} & \textbf{--} & \textbf{--} & \textbf{--} & \textbf{--} & \textbf{--} & \textbf{--} & \textbf{--} & \textbf{--} & \textbf{--} \\
TransFusion-LiDAR~\cite{bai2022transfusion} & CVPR 22 & 70.1 & 65.5 & 86.9 & 60.8 & 73.1 & 43.4 & 25.2 & 87.5 & 72.9 & 57.3 & 77.2 & 70.3 \\
DSVT~\cite{wang2023dsvt} & CVPR 23 & 71.1 & 66.4 & 87.4 & 62.6 & 75.9 & 42.1 & 25.3 & 88.2 & 74.8 & 58.7 & 77.9 & 71.0 \\
HEDNet~\cite{zhang2023hednet} & NeurIPS 23 & 71.4 & 66.7 & 87.7 & 60.6 & 77.8 & \textbf{50.7} & \textbf{28.9} & 87.1 & 74.3 & 56.8 & 76.3 & 66.9 \\
\midrule
\rowcolor[gray]{0.95}
{\ourMethod}-RetNet~(Ours) & \textbf{--} & 71.9 & 67.3 & 87.9 & 64.3 & \textbf{78.7} & 44.6 & 27.6 & 88.9 & 73.5 & 56.6 & 79.2 & \textbf{72.1}\\
\rowcolor[gray]{0.95}
{\ourMethod}-RWKV~(Ours) & \textbf{--} & 71.7 & 66.8 & \textbf{88.1} & 59.0 & 77.6 & 46.6 & 28.0 & \textbf{89.7} & 74.3 & 56.2 & 80.1 & 68.3\\
\rowcolor[gray]{0.95}
{\ourMethod}-Mamba~(Ours) & \textbf{--} & \textbf{72.1} & \textbf{68.0} & 87.9 & \textbf{64.9} & 77.6 & 44.4 & 28.5 & 89.6 & \textbf{75.6} & \textbf{59.4} & \textbf{80.8} & 71.6 \\
\midrule
\multicolumn{14}{c}{Performances on the \textit{test} set}\\
\midrule
TransFusion-LiDAR~\cite{bai2022transfusion} & CVPR 22 & 70.2 & 65.5 & 86.2 & 56.7 & 66.3 & 58.8 & 28.2 & 86.1 & 68.3 & 44.2 & 82.0 & 78.2\\
DSVT~\cite{wang2023dsvt} & CVPR 23 & 72.7 & 68.4 & 86.8 & 58.4 & 67.3 & 63.1 & \textbf{37.1} & 88.0 & 73.0 & 47.2 & 84.9 & 78.4 \\
HEDNet~\cite{zhang2023hednet} & NeurIPS 23 & 72.0 & 67.7 & 87.1 & 56.5 & \textbf{70.4} & 63.5 & 33.6 & 87.9 & 70.4 & 44.8 & 85.1 & 78.1 \\
\midrule
\rowcolor[gray]{0.95}
{\ourMethod}-Mamba~(Ours) & \textbf{--} & \textbf{73.9} & \textbf{69.8} & \textbf{87.2} & \textbf{61.1} & 68.9 & \textbf{65.0} & 36.3 & \textbf{90.0} & \textbf{74.0} & \textbf{49.2} & \textbf{87.3} & \textbf{79.5} \\
\bottomrule
\end{tabular}
 }
\label{tab:nuscenes_main_results}
\end{table}


\begin{table}[t!]
    \caption{Comparison with prior methods on Argoverse V2 \textit{validation} set. ‘Vehicle’, ‘C-Barrel’, ‘MPC-Sign’, ‘A-Bus’, ‘C-Cone’, ‘V-Trailer’, ‘MBT’, ‘W-Device’ and ‘W-Rider’ are short for regular vehicle, construction barrel, mobile pedestrian crossing sign, articulated bus, construction cone, vehicular trailer, message board trailer, wheeled device, and wheeled rider.}
    \small
    \centering
    \setlength{\tabcolsep}{1.1mm}{}
    \resizebox{1.0\linewidth}{!}{
    \begin{tabular}{l|c|rrrrrrrrrrrrrrrrrrrrrrrrrrrrrrrr}
      \toprule
      Method & \rotatebox{90}{mAP} & \rotatebox{90}{Vehicle} & \rotatebox{90}{Bus} & \rotatebox{90}{Pedestrian} & \rotatebox{90}{Stop Sign} &\rotatebox{90}{Box Truck} &\rotatebox{90}{Bollard} &\rotatebox{90}{C-Barrel} &\rotatebox{90}{Motorcyclist} &\rotatebox{90}{MPC-Sign} &\rotatebox{90}{Motorcycle} &\rotatebox{90}{Bicycle} &\rotatebox{90}{A-Bus} &\rotatebox{90}{School Bus} &\rotatebox{90}{Truck Cab} &\rotatebox{90}{C-Cone} &\rotatebox{90}{V-Trailer} &\rotatebox{90}{Sign} &\rotatebox{90}{Large Vehicle} &\rotatebox{90}{Stroller} &\rotatebox{90}{Bicyclist} &\rotatebox{90}{Truck} & \rotatebox{90}{MBT} &\rotatebox{90}{Dog} & \rotatebox{90}{Wheelchair} & \rotatebox{90}{W-Device}  & \rotatebox{90}{W-Rider} \\
      \midrule
      CenterPoint~\cite{yin2021center} & 22.0 & 67.6 & 38.9 & 46.5 & 16.9 & 37.4 & 40.1 & 32.2 & 28.6 & 27.4 & 33.4 & 24.5 & 8.7 & 25.8 & 22.6 & 29.5 & 22.4 & 6.3 & 3.9 & 0.5 & 20.1 & 22.1 & 0.0 & 3.9 & 0.5 & 10.9 & 4.2 \\
      HEDNet~\cite{zhang2023hednet} & 37.1 & 78.2 & 47.7 & 67.6 & 46.4 & 45.9 & 56.9 & 67.0 & 48.7 & 46.5 & 58.2 & 47.5 & 23.3 & 40.9 & 27.5 & 46.8 & 27.9 & 20.6 & 6.9 & 27.2 & 38.7 & 21.6 & 0.0 & 30.7 & 9.5 & 28.5 & 8.7 \\
      VoxelNeXt~\cite{chen2023voxelnext} & 30.7 & 72.7 & 38.8 & 63.2 & 40.2 & 40.1 & 53.9 & 64.9 & 44.7 & 39.4 & 42.4 & 40.6 & 20.1 & 25.2 & 19.9 & 44.9 & 20.9 & 14.9 & 6.8 & 15.7 & 32.4 & 16.9 & 0.0 & 14.4 & 0.1 & 17.4 & 6.6 \\
      FSDv1~\cite{fan2022fully} & 28.2 & 68.1 & 40.9 & 59.0 & 29.0 & 38.5 & 41.8 & 42.6 & 39.7 & 26.2 & 49.0 & 38.6 & 20.4 & 30.5 & 14.8 & 41.2 & 26.9 & 11.9 & 5.9 & 13.8 & 33.4 & 21.1 & 0.0 & 9.5 & 7.1 & 14.0 & 9.2 \\
      FSDv2~\cite{fan2023fsd} & 37.6 & \textbf{77.0} & 47.6 & 70.5 & 43.6 & 41.5 & 53.9 & 58.5 & 56.8 & 39.0 & 60.7 & 49.4 & 28.4 & 41.9 & \textbf{30.2} & 44.9 & \textbf{33.4} & 16.6 & 7.3 & 32.5 & 45.9 & \textbf{24.0} & \textbf{1.0} & 12.6 & \textbf{17.1} & 26.3 & 17.2 \\
      SAFDNet~\cite{zhang2024safdnet}& 39.7 & 78.5 & \textbf{49.4} & 70.7 & 51.5 & 44.7 & 65.7 & 72.3 & 54.3 & \textbf{49.7} & 60.8 & 50.0 & \textbf{31.3} & \textbf{44.9} & 24.7 & 55.4 & 31.4 & 22.1 & 7.1 & 31.1 & 42.7 & 23.6 & 0.0 & 26.1 & 1.4 & 30.2 & 11.5 &  \\
      \midrule
      \rowcolor[gray]{0.95}
      {\ourMethod}-RetNet& 40.7 & 74.7 & 41.0 & 72.7 & 47.5 & 44.2 & 66.9 & \textbf{77.0} & 57.1 & 48.3 & 63.7 & 55.1 & 27.0 & 42.5 & 25.2 & 57.9 & 29.7 & 22.0 & 6.9 & 39.3 & 47.3 & 19.9 & 0.0 & 28.8 & 12.8 & \textbf{37.7} & \textbf{12.6} &  \\
      \rowcolor[gray]{0.95}
      {\ourMethod}-RWKV & 41.1 & 76.3 & 44.6 & \textbf{74.0} & 52.1 & \textbf{46.0} & \textbf{68.1} & 75.8 & 55.8 & 49.4 & 62.8 & 55.3 & 27.1 & 42.9 & 25.9 & \textbf{60.1} & 30.9 & 22.2 & \textbf{9.3} & 36.5 & \textbf{55.3} & 23.2 & 0.0 & 27.8 & 7.1 & 37.6 & 11.4 &  \\
      \rowcolor[gray]{0.95}
      {\ourMethod}-Mamba& \textbf{41.5} & 75.1 & 43.6 & 73.9 & \textbf{53.9} & 45.1 & 66.4 & 74.7 & \textbf{61.3} & 48.7 & \textbf{65.1} & \textbf{56.2} & 21.7 & 42.7 & 25.3 & 58.4 & 28.9 & \textbf{23.6} & 8.3 & \textbf{49.5} & 47.3 & 19.0 & 0.0 & \textbf{31.4} & 8.7 & 37.6 & 11.8 &  \\
       \bottomrule
       \end{tabular}}
\label{tab:argo_results}    
\end{table}

\begin{table}[t!]
\caption{Comparison with previous methods on ONCE \textit{validation} set. We use the center head of CenterPoint for a fair comparison.}
\vspace{-5pt}
\centering
\resizebox{1.0\linewidth}{!}{
\begin{tabular}{l|cccc|cccc|cccc|c}
        \toprule
        \multirow{2}{*}{Method} & \multicolumn{4}{c|}{ Vehicle} & \multicolumn{4}{c|}{Pedestrian} & \multicolumn{4}{c|}{ Cyclist} & \multirow{2}{*}{mAP} \\ 
        & overall & 0-30m & 30-50m & 50m-inf & overall & 0-30m & 30-50m & 50m-inf & overall & 0-30m & 30-50m & 50m-inf &  \\ 
        \midrule
        PointRCNN~\cite{shi2019pointrcnn} & 52.1 & 74.5 & 40.9 & 16.8 & 4.3 & 6.2 & 2.4 & 0.9 & 29.8 & 46.0 & 20.9 & 5.5 & 28.7 \\ 
        PointPillars~\cite{lang2019pointpillars} & 68.6 & 80.9 & 62.1 & 47.0 & 17.6 & 19.7 & 15.2 & 10.2 & 46.8 & 58.3 & 40.3 & 25.9 & 44.3 \\
        SECOND~\cite{yan2018second} & 71.2 & 84.0 & 63.0 & 47.3 & 26.4 & 29.3 & 24.1 & 18.1 & 58.0 & 70.0 & 52.4 & 34.6 & 51.9 \\ 
        PV-RCNN~\cite{shi2020pv} & 77.8 & \textbf{89.4} & \textbf{72.6} & 58.6 & 23.5 & 25.6 & 22.8 & 17.3 & 59.4 & 71.7 & 52.6 & 36.2 & 53.6 \\ 
        CenterPoint~\cite{yin2021center} & 66.8 & 80.1 & 59.6 & 43.4 & 49.9 & 56.2 & 42.6 & \textbf{26.3} & 63.5 & 74.3 & 57.9 & 41.5 & 60.1 \\ 
        PointPainting~\cite{vora2020pointpainting} & 66.2 & 80.3 & 59.8 & 42.3 & 44.8 & 52.6 & 36.6 & 22.5 & 62.3 & 73.6 & 57.2 & 40.4 & 57.8 \\ 
        \midrule
        \rowcolor[gray]{0.95}
        {\ourMethod}-RetNet & 78.1 & 88.7 & 72.4 & \textbf{58.5} & 52.4 & 60.5 & 43.6 & \textbf{26.3} & 68.3 & \textbf{79.4} & 62.9 & 46.1 & 66.3 \\
        \rowcolor[gray]{0.95}
        {\ourMethod}-RWKV & \textbf{78.3} & 89.2 & \textbf{72.6} & 56.7 & 50.6 & 60.0 & 40.4 & 24.2 & 68.4 & \textbf{79.4} & \textbf{63.2} & 45.7 & 65.8 \\
        \rowcolor[gray]{0.95}
        {\ourMethod}-Mamba & 78.2 & 89.1 & \textbf{72.6} & 57.5 & \textbf{53.2} & \textbf{62.4} & \textbf{44.0} & 24.5 & \textbf{68.5} & 79.2 & \textbf{63.2} & \textbf{47.1} & \textbf{66.6} \\
        \bottomrule
        
    \end{tabular}
 }
\label{tab:once_results}
\vspace{-5pt}
\end{table}

\noindent\textbf{Results on nuScenes.}
We also evaluate our {\ourMethod} on nuScenes \textit{validation} and \textit{test} set~\cite{caesar2020nuscenes} further to verify the effectiveness of our {\ourMethod}. As shown in Table~\ref{tab:nuscenes_main_results}, on nuScenes \textit{validation} set, our {\ourMethod}-RetNet, {\ourMethod}-RWKV, and {\ourMethod}-Mamba achieves 71.9, 71.7, and 72.1 NDS, respectively, which outperforms the previous advanced methods DSVT~\cite{wang2023dsvt} and HEDNet~\cite{zhang2023hednet}. Besides, our {\ourMethod}-Mamba even brings a new SOTA on nuScenes \textit{test} benchmark, which beats the previous advanced method DSVT with 1.2 NDS and 1.4 mAP, effectively illustrating the superiority of our {\ourMethod}. Note that all results of our {\ourMethod} are conducted without any test-time augmentation and model ensembling.

\noindent\textbf{Results on Argoverse V2.} To further verify the effectiveness of our {\ourMethod} on the long-range perception, we evaluate the experiments on Argoverse V2 \textit{validation} set. For a fair comparison, we adopt the same detection head~\cite{yin2021center} with VoxelNext~\cite{chen2023voxelnext} and SAFDNet~\cite{zhang2024safdnet} for long-range perception.
As shown in Table~\ref{tab:argo_results}, our {\ourMethod}-RetNet, {\ourMethod}-RWKV, and {\ourMethod}-Mamba achieve the detection performance with 40.7 mAP, 41.1 mAP and 41.5 mAP, all three of which have outperformed the previous SOTA method SAFDNet~\cite{zhang2024safdnet}, leading to new SOTA results. These superior results clearly illustrate the effectiveness of our {\ourMethod}.


\noindent\textbf{Results on ONCE.} We also evaluated our {\ourMethod} on ONCE \textit{validation} set to further verify the effectiveness of our {\ourMethod}. As shown in Table~\ref{tab:once_results}, our {\ourMethod}-RetNet, {\ourMethod}-RWKV, and {\ourMethod}-Mamba produces advanced detection performance with 66.3 mAP, 65.8 mAP, and 66.6 mAP, respectively. It is worth mentioning that our {\ourMethod}-Mamba outperforms the previous SOTA method CenterPoint~\cite{yin2021center} with 6.5 mAP, leading to a new SOTA result. These experiments illustrate the superiority of our {\ourMethod}.


\subsection{Ablation Study}
In this section, we conduct ablation studies of {\ourMethod} on the WOD \textit{validation} set with 20\% training data. If not specified, we adopt {\ourMethod}-Mamba as our default model and train our model with 12 epochs in the following ablation studies. For more experiments, please refer to our appendix.


\begin{table}[t!]
\caption{Ablation study for each component in {\ourMethod}. Here, the large group size means that we set it as (4096, 2048, 1024, 512) for four blocks~(also refer to the section of our implementation details), otherwise, we set a small group size as (256, 256, 256, 256).}
\vspace{-5pt}
\small
\centering
\setlength{\tabcolsep}{8pt}
\resizebox{1.0\linewidth}{!}{
\begin{tabular}{c|c|c|c|c|c|c}
\toprule
\multicolumn{1}{c|}{\multirow{2}{*}{Large Group Size}} & \multirow{2}{*}{3D Spatial Feature Descriptor} & \multirow{2}{*}{Voxel Generation} & \multicolumn{3}{c|}{3D AP/APH~(L2)} & \multirow{2}{*}{\shortstack{mAP/mAPH \\ (L2)}}\\
 & &  & \emph{Vehicle} & \emph{Pedestrian} & \emph{Cyclist}\\
\midrule
\textbf{--}  & \textbf{--}  & \textbf{--}  & 65.6/65.2 & 72.3/65.0  & 68.3/67.2 & 68.8/65.8  \\
$\checkmark$ & \textbf{--}  & \textbf{--}  & 66.2/65.7 & 73.7/67.2  & 68.7/67.6  &69.5/66.9  \\
$\checkmark$  & $\checkmark$  & \textbf{--}  & 66.5/66.1 & 74.8/69.6  & 70.9/70.0 &70.8/68.6 \\
$\checkmark$  & \textbf{--}  & $\checkmark$ & 66.4/66.0 &73.5/67.4  & 70.4/69.3 &70.1/67.6  \\
$\checkmark$  & $\checkmark$  & $\checkmark$  & \textbf{67.0/66.6} & \textbf{75.4/70.2} & \textbf{71.9/71.0}  & \textbf{71.4/69.3}  \\
\bottomrule
\end{tabular}
 }
\label{tab:ab_com}
\vspace{-10pt}
\end{table}

\noindent\textbf{Ablation Study of {\ourMethod}.}
To illustrate the effectiveness of our proposed {\ourMethod}, we conduct the ablation study for each component, including the design of large group size, 3D spatial feature descriptor, and voxel generation in Table~\ref{tab:ab_com}. Here, our baseline is proposed {\ourMethod} that removes the design of large group size, 3D spatial feature descriptor, and voxel generation.
In Table~\ref{tab:ab_com}, we observe that the design of large group size even brings 1.1 mAPH/L2 performance improvement, which illustrates the benefits of performing long-range feature interaction with the help of linear RNN. Then, we integrate the 3D spatial feature descriptor, which further produces an obvious performance improvement with 1.7 mAPH/L2. This demonstrates the superiority of the 3D spatial feature descriptor in compensating for the lack of capturing spatial information of linear RNNs. Furthermore, we notice that the 
3D spatial feature descriptor is very helpful to small objects~(\textit{e.g.}, Pedestrians) thanks to its capability of extracting the local information of 3D objects. To address the challenge of feature representation in highly sparse point clouds, we adopt voxel generation to enhance the features of foregrounds, which brings a promising gain of 0.7 mAPH/L2~(67.6 \textit{vs.} 66.9). By combining all components, our {\ourMethod} achieves a superior performance of 69.3 mAPH/L2, which outperforms the baseline of 3.5 mAPH/L2.


\noindent\textbf{Superiority of 3D Spatial Feature Descriptor.} To further verify the necessity of 3D spatial feature descriptor, we provide the comparison with two available manners including the MLP and linear RNN to replace our descriptor in Table~\ref{tab:sfd}.
Here, we set our {\ourMethod} without 3D spatial feature descriptor as the baseline in this part. We observe that MLP even does not bring promising performance improvement in terms of mAPH/L2 since MLP lacks the ability to capture local 3D spatial information.
Furthermore, considering the limited receptive field of MLP, we adopt a linear group RNN operator to replace MLP. We find that there is only slight performance improvement with 0.3 mAPH/L2, which indicates that the linear group RNN might not be good at modeling local spatial relationships although it has the strong capability to establish long-range relationships.
In contrast, our 3D spatial feature descriptor brings obvious performance improvement, which boosts the baseline of 1.7 mAPH/L2. This effectively illustrates the superiority of the 3D spatial feature descriptor in compensating for the lack of local 3D spatial-aware modeling in the linear group RNN.

\begin{table}[t!]
\caption{Ablation study for 3D Spatial Feature Descriptor~(3D SFD) in {\ourMethod}.}
\vspace{-5pt}
\small
\centering
\setlength{\tabcolsep}{22pt}
\resizebox{1.0\linewidth}{!}{
\begin{tabular}{l|c|c|c|c}
\toprule
\multicolumn{1}{l|}{\multirow{2}{*}{Methods}} & \multicolumn{3}{c|}{3D AP/APH~(L2)} & \multirow{2}{*}{\shortstack{mAP/mAPH \\ (L2)}}\\
& \emph{Vehicle} & \emph{Pedestrian} & \emph{Cyclist}\\
\midrule
Baseline    &66.4/66.0 &73.5/67.4  & 70.4/69.3 &70.1/67.6 \\
MLP    &66.6/66.2 & 74.1/68.1 &70.0/69.0  &70.2/67.7 \\
Linear Group RNN    & 66.4/66.0 & 74.0/68.2 &70.5/69.5  &70.3/67.9 \\
3D SFD~(Ours)   & \textbf{67.0/66.6} & \textbf{75.4/70.2} & \textbf{71.9/71.0}  & \textbf{71.4/69.3}  \\
\bottomrule
\end{tabular}
 }
\label{tab:sfd}
\vspace{-5pt}
\end{table}

\begin{table}[t!]
\caption{Ablation study for voxel generation in {\ourMethod}. ``Baseline'' indicates no voxel generation. ``Zero Feats'' and ``K-NN Feats'' indicate initializing features to all zeros and K-NN features, respectively.
``Auto-Regressive'' uses the {\ourMethod} block based on linear group RNN for its auto-regressive property. ``Sparse-Conv'' maintains the same structure as the {\ourMethod} block but replaces the linear group RNN with 3D sub-manifold convolution.}
\vspace{-5pt}
\small
\centering
\setlength{\tabcolsep}{12pt}
\resizebox{1.0\linewidth}{!}{
\begin{tabular}{c|c|c|c|c|c}
\toprule
\multicolumn{1}{c|}{\multirow{2}{*}{Index}} & \multicolumn{1}{c|}{\multirow{2}{*}{Methods}} & \multicolumn{3}{c|}{3D AP/APH~(L2)} & \multirow{2}{*}{\shortstack{mAP/mAPH \\ (L2)}}\\
& & \emph{Vehicle} & \emph{Pedestrian} & \emph{Cyclist}\\
\midrule
\uppercase\expandafter{\romannumeral1}  & Baseline   & 66.5/66.1 & 74.8/69.6  & 70.9/70.0 &70.8/68.6 \\
\uppercase\expandafter{\romannumeral2}  & Zero Feats + Sparse-Conv  &64.6/64.2 & 72.8/67.4 &69.3/68.3  &68.9/66.6 \\
\uppercase\expandafter{\romannumeral3}  & K-NN Feats + Auto-Regressive   &66.5/66.1 & 74.0/68.7 &71.1/70.1  &70.5/68.3 \\
\uppercase\expandafter{\romannumeral4}  & Zero Feats + Auto-Regressive~(Ours) & \textbf{67.0/66.6} & \textbf{75.4/70.2} & \textbf{71.9/71.0}  & \textbf{71.4/69.3}  \\
\bottomrule
\end{tabular}
 }
\label{tab:vfg}
\vspace{-15pt}
\end{table}

\noindent\textbf{Effectiveness of Voxel Generation.} Voxel generation is applied to enhance the feature representation of objects in highly sparse point clouds for accurate 3D object detection. 
Therefore, to explore the effectiveness of our proposed voxel generation, we present the comparison with several available manners in Table~\ref{tab:vfg}. First, we compare our results of \uppercase\expandafter{\romannumeral4} with 
{II} by only replacing the operator of linear group RNN in {\ourMethod} block with 3D sub-manifold convolution to generate the diffused features. We find that the manner of \uppercase\expandafter{\romannumeral4}~(69.3 \textit{vs.} 66.6) significantly outperforms the performance of \uppercase\expandafter{\romannumeral2} in terms of mAPH/L2. This benefits from the linear group RNN's ability to model long-range feature interactions, generating a more reliable feature representation through its auto-regressive capacity, demonstrating the superiority of voxel generation with the linear group RNN.
To further illustrate that the effectiveness of voxel generation is from its auto-regressive property of {\ourMethod} block rather than a strong feature extractor, we initialize the diffused features of the foreground voxels by K-NN operation~(\uppercase\expandafter{\romannumeral3}) instead of the manner of all-zeros features~(\uppercase\expandafter{\romannumeral4}) and then feed them to the same following {\ourMethod} block for voxel generation. In Table~\ref{tab:vfg}, we find that the manner of \uppercase\expandafter{\romannumeral3} is inferior to \uppercase\expandafter{\romannumeral4} by 1.0 mAPH/L2. This clearly illustrates that our voxel generation is benefiting from its auto-regressive property of {\ourMethod} block. Finally, compared with the baseline (\uppercase\expandafter{\romannumeral1}), our voxel generation (\uppercase\expandafter{\romannumeral4}) can obtain a promising performance improvement, which verifies its effectiveness.


\section{Conclusion}
In this paper, we have presented a simple and effective window-based framework termed {\ourMethod}, which can capture the long-range relationship by adopting linear RNN for large groups. 
Specifically, {\ourMethod} incorporates a proposed {\ourMethod} block to unlock the great potential of linear RNNs in modeling a long-range relationship and a voxel generation strategy to obtain more discriminative feature representation in sparse point clouds. 
Extensive ablation studies demonstrate the effectiveness of our proposed components.
Additionally, the generalization of our {\ourMethod} is verified by performing different linear group RNN operators.
Benefiting from our well-designed framework and the proposed superior components, our {\ourMethod}-Mamba achieves state-of-the-art performance on the challenging Waymo and nuScenes datasets.

\textbf{Limitations.} Although our {\ourMethod} based on the linear group RNN can perform long-range feature interaction with linear complexity, the corresponding running speed still needs further improvement since linear RNNs are not as efficient as transformers in parallel computing.

{\small
\bibliographystyle{plain}
\bibliography{ref}
}

\newpage
\appendix

\section{Appendix}
The appendix is organized as follows.
First, in section~\ref{sec:kitti}, we provide more types of linear RNN operators~(\textit{e.g.}, RetNet, RWKV, Mamba, xLSTM, and TTT) based on our {\ourMethod} framework for 3D detection on a small but popular dataset KITTI for a quick experience.
Second, we present extra experiments on the WOD~\cite{sun2020scalability} \textit{validation} set in section~\ref{sec:exps}, including the placement of the 3D spatial feature descriptor, the impact of different window sizes and different group sizes in inference, and the ratio $r$ in voxel generation.
Third, we provide the comparison of computation cost and parameter size in section~\ref{sec:cmp_cost}, and detailed information of {\ourMethod} structure in section~\ref{sec:arch}. 
Forth, in section~\ref{sec:vis_feature}, we visualize the feature maps of different {\ourMethod} blocks to illustrate the rationality of distinguishing foreground voxels based on feature response. 
Finally, we provide the comparison of qualitative results with DSVT~\cite{wang2023dsvt} and qualitative results of {\ourMethod} to demonstrate the superiority of our {\ourMethod} in section~\ref{sec:cmp_results} and section~\ref{sec:qualitative}. Besides, we discuss the broader impacts in section~\ref{sec:broader}.

\subsection{Experiments on KITTI dataset}
\label{sec:kitti}

\begin{table*}[h]
\caption{Effectiveness on the KITTI \textit{validation} set for \emph{Car}, \emph{Pedestrian}, and \emph{Cyclist}. * represents our reproduced results by keeping the same configures except for their 3D backbones for a fair comparison. Our {\ourMethod} supports different representative linear RNN operators (TTT, xLSTM, RetNet, RWKV, and Mamba). mAP is calculated by all categories and all difficulties with recall 11.}
\small
\centering
\resizebox{1.0\linewidth}{!}{
\begin{tabular}{l|c|c|c|c|c|c|c|c|c|c}
\toprule
\multirow{2}{*}{Method} 
&\multicolumn{3}{c|}{\emph{Car}} 
& \multicolumn{3}{c|}{\emph{Pedestrian}} 
& \multicolumn{3}{c|}{\emph{Cyclist}} 
& \multicolumn{1}{c}{\multirow{2}{*}{mAP}} \\
& Easy & Moderate & Hard & Easy & Moderate & Hard & Easy & Moderate & Hard & \multicolumn{1}{c}{}  \\
\midrule
VoxelNet~\cite{zhou2018voxelnet} &77.5 &65.1 &57.7 &39.5 &33.7 &31.5 &61.2 &48.4 &44.4  &51.0 \\
SECOND~\cite{yan2018second} &83.1 &73.7 &66.2 &51.1 &42.6 &37.3 &70.5 &53.9 &46.9  &58.4  \\
PointPillars~\cite{lang2019pointpillars} & 79.1 &75.0 &68.3 &52.1 &43.5&41.5& 75.8 & 59.1 & 52.9  &60.8 \\
PointRCNN~\cite{shi2019pointrcnn}  & 85.9 & 75.8 & 68.3 & 49.4 &41.8 &38.6 & 73.9 &59.6 & 53.6 & 60.8 \\
TANet~\cite{liu2020tanet} &83.8 &75.4 &67.7 & 54.9 & 46.7 & 42.4 & 73.8 & 59.9 & 53.5  & 62.0  \\
DSVT-Pillar*~\cite{wang2023dsvt} & 87.3 & 77.4 & 76.2 & 61.4 & 56.8 & 51.8 & 82.3 & 67.1 & 63.7  & 69.3 \\
DSVT-Voxel*~\cite{wang2023dsvt} & 87.8 & 77.8 & 76.8 & 66.1 & 59.7 & 55.2 & 83.5 & 66.7 & 63.2 & 70.8 \\
\midrule
\rowcolor[gray]{0.95}
{\ourMethod}-TTT & 87.9 & 78.0 & 76.7 & 63.4 & 58.6 & 53.7 & 84.0 & 69.6 & 64.5 & 70.7 \\
\rowcolor[gray]{0.95}
{\ourMethod}-xLSTM & 87.7 & 77.9 & 76.8 & 66.6 & 59.3 & 54.0 & 82.4 & 67.4 & 63.4 & 70.6 \\
\rowcolor[gray]{0.95}
{\ourMethod}-RetNet & 88.0 & 77.9 & 76.7 & 67.4 & 60.2 & 55.8 & 83.6 & 69.6 & 64.6 & 71.5 \\
\rowcolor[gray]{0.95}
{\ourMethod}-Mamba & \textbf{88.6} & \textbf{78.3} & 77.2 & 67.2 & 60.2 & 55.6 & 83.0 & 68.6 & 63.9 & 71.4 \\
\rowcolor[gray]{0.95}
{\ourMethod}-RWKV & 88.5 & \textbf{78.3} & \textbf{77.1} & \textbf{68.9} & \textbf{62.2} & \textbf{58.1} & \textbf{89.6} & \textbf{71.2} & \textbf{66.9} & \textbf{73.4} \\
\bottomrule
\end{tabular}
}
\label{kitti}
\end{table*}

\textbf{KITTI Dataset.}
KITTI~\cite{geiger2012we} is a popular benchmark dataset for autonomous driving, which consists of 7481 training frames and 7518 test frames for 3D object detection. We follow the dataset splitting protocol in \cite{qi2018frustum} and further split the 7481 training frames into 3712 frames for \textit{training} set and 3769 frames for \textit{validation} set. For the 3D detection task, KITTI dataset mainly detects Car, Pedestrian, and Cyclist for three difficulty levels, \textit{i.e.}, Easy, Moderate, and Hard. And the mean Average Precision~(mAP) using 11 recall positions is adopted as the evaluation metric.

\noindent\textbf{Results on KITTI.}
We conduct experiments on the KITTI \textit{validation} set to illustrate the generalization of {\ourMethod} for different linear RNN operators. We select some representative linear RNN operators (TTT~\cite{sun2024learning}, xLSTM~\cite{beck2024xlstm}, RetNet~\cite{sun2023retentive}, Mamba~\cite{gu2023mamba}, and RWKV~\cite{peng2023rwkv}) for {\ourMethod}. We adopt the same training parameters~(\textit{i.e.}, number of epochs, learning rate, optimizer) with SECOND~\cite{yan2018second}. Besides, we use the same BEV backbone and the detection head with SECOND~\cite{yan2018second}.
For a fair comparison, we keep the same configure of DSVT~\cite{wang2023dsvt} and all our {\ourMethod} methods except 3D backbones. 
As shown in Table~\ref{kitti}, {\ourMethod}-RetNet, {\ourMethod}-Mamba, and {\ourMethod}-RWKV outperforms DSVT-Voxel by 0.7 mAP, 0.6 mAP, and 2.6 mAP. These experiments demonstrate the generalization and effectiveness of our linear RNN-based framework {\ourMethod}.

\subsection{Extra Experiments}
\label{sec:exps}

\begin{table}[!ht]
\caption{The Placement of 3D Spatial Feature Descriptor.}
\small
\centering
\setlength{\tabcolsep}{20pt}
\resizebox{1.0\linewidth}{!}{
\begin{tabular}{l|c|c|c|c}
\toprule
\multicolumn{1}{l|}{\multirow{2}{*}{Methods}} & \multicolumn{3}{c|}{3D AP/APH~(L2)} & \multirow{2}{*}{\shortstack{mAP/mAPH \\ (L2)}}\\
& \emph{Vehicle} & \emph{Pedestrian} & \emph{Cyclist}\\
\midrule
Baseline & 66.4/66.0 &73.5/67.4  & 70.4/69.3 &70.1/67.6 \\
Placement 1  &66.5/66.1 &74.8/69.1  & 71.1/70.2 & 70.1/68.6 \\
Placement 2 (Ours) & \textbf{67.0/66.6} & \textbf{75.4/70.2} & \textbf{71.9/71.0}  & \textbf{71.4/69.3}  \\
\bottomrule
\end{tabular}
 }
\label{tab:sfd_pos}
\end{table}

\begin{table}[!ht]
\caption{The ratio $r$ in voxel genenration.}
\small
\centering
\setlength{\tabcolsep}{20pt}
\resizebox{1.0\linewidth}{!}{
\begin{tabular}{l|c|c|c|c}
\toprule
\multicolumn{1}{l|}{\multirow{2}{*}{Ratio}} & \multicolumn{3}{c|}{3D AP/APH~(L2)} & \multirow{2}{*}{\shortstack{mAP/mAPH \\ (L2)}}\\
& \emph{Vehicle} & \emph{Pedestrian} & \emph{Cyclist}\\
\midrule
$0$ & 66.5/66.1 & 74.8/69.6  & 70.9/70.0 &70.8/68.6 \\
$0.2$ (Ours) & 67.0/66.6 & \textbf{75.4}/\textbf{70.2} & 71.9/71.0  & 71.4/\textbf{69.3}  \\
$0.5$  & \textbf{67.2}/\textbf{66.8} & 75.3/70.0 & \textbf{72.1}/\textbf{71.1} & \textbf{71.5}/\textbf{69.3} \\
\bottomrule
\end{tabular}
 }
\label{tab:rate}
\end{table}

\begin{table}[htbp]
\caption{Comparison of different window and group sizes in inference on WOD \textit{validation} set~(train with 100\% training data). \textbf{Bold} denotes the result of {\ourMethod} with the default settings in the main paper.}
\setlength{\tabcolsep}{12pt}
\centering
\resizebox{1.0\linewidth}{!}{
    \centering
    \begin{tabular}{c|c|c}
    \toprule
    Window Size &Group Size  & mAP/mAPH (L2) \\ \midrule
    {$[7, 7, 32]$, $[7, 7, 16]$, $[7, 7, 8]$, $[7, 7, 4]$} &{[4096, 2048, 1024, 512]}  & 73.24 \\ \midrule
    {$[13, 13, 32]$, $[13, 13, 16]$, $[13, 13, 8]$, $[13, 13, 4]$} &{[4096, 2048, 1024, 512]}  & \textbf{73.24} \\ \midrule
    {$[25, 25, 32]$, $[25, 25, 16]$, $[25, 25, 8]$, $[25, 25, 4]$} &{[4096, 2048, 1024, 512]}  & 73.25 \\ 
    \midrule
    \midrule
    {$[13, 13, 32]$, $[13, 13, 16]$, $[13, 13, 8]$, $[13, 13, 4]$} &{[2048, 1024, 512, 256]}  & 73.18 \\ \midrule
    {$[13, 13, 32]$, $[13, 13, 16]$, $[13, 13, 8]$, $[13, 13, 4]$} &{[4096, 2048, 1024, 512]}  & \textbf{73.24} \\ \midrule
    {$[13, 13, 32]$, $[13, 13, 16]$, $[13, 13, 8]$, $[13, 13, 4]$} &{[8192, 4096, 2048, 1024]}  & 73.02 \\ 
    \bottomrule
    \end{tabular}
    }
\label{tab:comparison}
\end{table}

\noindent\textbf{The Placement of 3D Spatial Feature Descriptor.}
We conduct experiments about the placement of the 3D spatial feature descriptor, as shown in Table~\ref{tab:sfd_pos}. We regard the manner that does not adopt the 3D SFD of our {\ourMethod} as the baseline.
Here, we provide two available manners: Placement 1 and Placement 2. 
For Placement 1, we place the 3D SFD after voxel merging. For Placement 2, we place 3D SFD before the voxel merging. Compared to the baseline, Placement 1 brings 1.0 mAPH/L2 improvement, which demonstrates the effectiveness of 3D SFD in compensating for the lack of local 3D spatial-aware modeling in linear RNNs. Moreover, Placement 2 further brings 0.7 mAPH/L2 improvement over Placement 1, which demonstrates the effectiveness of 3D SFD for reducing spatial information loss in the process of voxel merging.

\noindent\textbf{The Ratio in Voxel Generation.} We conduct the ablation study for foreground selection ratio $r$ in voxel generation. As shown in Table~\ref{tab:rate}, compared with baseline ($r=0$), the manner of setting $r=0.2$ brings 0.7 mAPH/L2 performance improvement. When we set a larger ratio $r=0.5$, the performance is improved slightly. Therefore, we set $r=0.2$ to balance the performance and computation cost.

\noindent\textbf{Different Window Sizes and Group Sizes in Inference.}
To analyze the impact of window size and group size in the inference process, 
we evaluate the results under the cases of different window sizes and group sizes with the same trained model of {\ourMethod}-Mamba~(\textit{i.e.}, window size=\{$(13, 13, 32)$, $(13, 13, 16)$, $(13, 13, 8)$, and $(13, 13, 4)$\} and group size=\{$4096$, $2048$, $1024$, $512$\}) on WOD 100\% training data. As shown in Table~\ref{tab:comparison}, surprisingly, we find that using different window sizes or group sizes during inference still does not significantly affect performance. This indicates that {\ourMethod} might decrease the strong dependence on hand-crafted priors and have good extrapolation ability.


\subsection{Comparison of Computation Cost and Parameter Size}
\label{sec:cmp_cost}
\begin{table}[htbp]
\caption{Comparison of computation cost and parameter size of different 3D backbones on the WOD \textit{validation} set.}
\small
\setlength{\tabcolsep}{12pt}
\resizebox{1.0\linewidth}{!}{
\centering
\begin{tabular}{l|c|c|c|c}
\toprule
Backbone & Operator & mAP/mAPH~(L2) & FLOPs~(G) & Params~(M) \\
\midrule
ResBackbone8$\times$~\cite{yin2021center} & Sparse Conv & 68.2/65.8 & 48.2 & 2.7 \\
\midrule
SST~\cite{fan2022embracing} & Transformer & 67.8/64.6 & 86.2 & 1.6 \\
FlatFormer~\cite{liu2023flatformer} & Transformer & 69.7/67.2 & 48.3 & 1.1 \\
DSVT-Pillar~\cite{wang2023dsvt} & Transformer & 73.2/71.0 & 88.2 & 2.7 \\
DSVT-Voxel~\cite{wang2023dsvt} & Transformer & 74.0/72.1 & 100.8 & 2.7 \\
\midrule
{\ourMethod}~(Ours) & RWKV & 74.7/72.8 & 79.4 & 3.0 \\
{\ourMethod}~(Ours) & RetNet & 74.6/72.7 & 51.3 & 2.0 \\
{\ourMethod}~(Ours) & Mamba & 75.1/73.2 & 58.5 & 1.4 \\
\bottomrule
\end{tabular}
 }
\label{tab:cost}
\end{table}

We compare different 3D backbones on the WOD \textit{validation} set. As shown in Table~\ref{tab:cost}, {\ourMethod} with RWKV~\cite{peng2024eagle}, RetNet~\cite{sun2023retentive}, and Mamba~\cite{gu2023mamba}, achieve 72.8, 72.7, and 73.2 mAPH/L2, respectively. Compared with other sparse convolution and transformer-based backbones, {\ourMethod} achieves superior accuracy while maintaining a satisfactory computation cost and parameter size. 

\subsection{Architecture Specifications}
\label{sec:arch}

\begin{table*}[htbp]
\caption{Detailed architecture specifications on Waymo Open dataset.}
\small
\renewcommand{\arraystretch}{1.5}
\setlength{\tabcolsep}{12pt}
\resizebox{1.0\linewidth}{!}{
\centering
\begin{tabular}{>{\centering\arraybackslash}m{1cm}|c|c|c|c}
\toprule
 & {\ourMethod}-RWKV & {\ourMethod}-RetNet & {\ourMethod}-Mamba & {\ourMethod}-Mamba-L \\
\toprule
\multirow{2}{*}{Block}& \makecell{Window Shape} & \makecell{Window Shape} & \makecell{Window Shape} & \makecell{Window Shape} \\
 & \makecell{Dim, Group Size} & \makecell{Dim, Group Size} & \makecell{Dim, Group Size} & \makecell{Dim, Group Size} \\
\midrule[\heavyrulewidth]
\multirow{2}{*}{Block 1} & \makecell{$[13, 13, 32]$} & \makecell{$[13, 13, 32]$} & \makecell{$[13, 13, 32]$} & \makecell{$[13, 13, 32]$} \\
 & \makecell{64, 4096} & \makecell{64, 4096} & \makecell{64, 4096} & \makecell{128, 4096} \\
\midrule
\multirow{2}{*}{Block 2} & \makecell{$[13, 13, 16]$} & \makecell{$[13, 13, 16]$} & \makecell{$[13, 13, 16]$} & \makecell{$[13, 13, 16]$} \\
 & \makecell{64, 2048} & \makecell{64, 2048} & \makecell{64, 2048} & \makecell{128, 2048} \\
\midrule
\multirow{2}{*}{Block 3} & \makecell{$[13, 13, 8]$} & \makecell{$[13, 13, 8]$} & \makecell{$[13, 13, 8]$} & \makecell{$[13, 13, 8]$} \\
 & \makecell{64, 1024} & \makecell{64, 1024} & \makecell{64, 1024} & \makecell{128, 1024} \\
\midrule
\multirow{2}{*}{Block 4} & \makecell{$[13, 13, 4]$} & \makecell{$[13, 13, 4]$} & \makecell{$[13, 13, 4]$} & \makecell{$[13, 13, 4]$} \\
 & \makecell{64, 512} & \makecell{64, 512} & \makecell{64, 512} & \makecell{128, 512} \\
\bottomrule
\end{tabular}
}
\normalsize
\label{table:arch-spec}
\end{table*}
As shown in Table~\ref{table:arch-spec}, the architecture specifications of the {\ourMethod} models ({\ourMethod}-RWKV, {\ourMethod}-RetNet, {\ourMethod}-Mamba, and {\ourMethod}-Mamba-L) on Waymo Open dataset are detailed in terms of window shape, dimension, and group size. For {\ourMethod}-Mamba-L, we set the dimension to 128 to double the channel of {\ourMethod}.

\begin{figure}[htbp]
    \centering
    \includegraphics[width=0.99\linewidth]{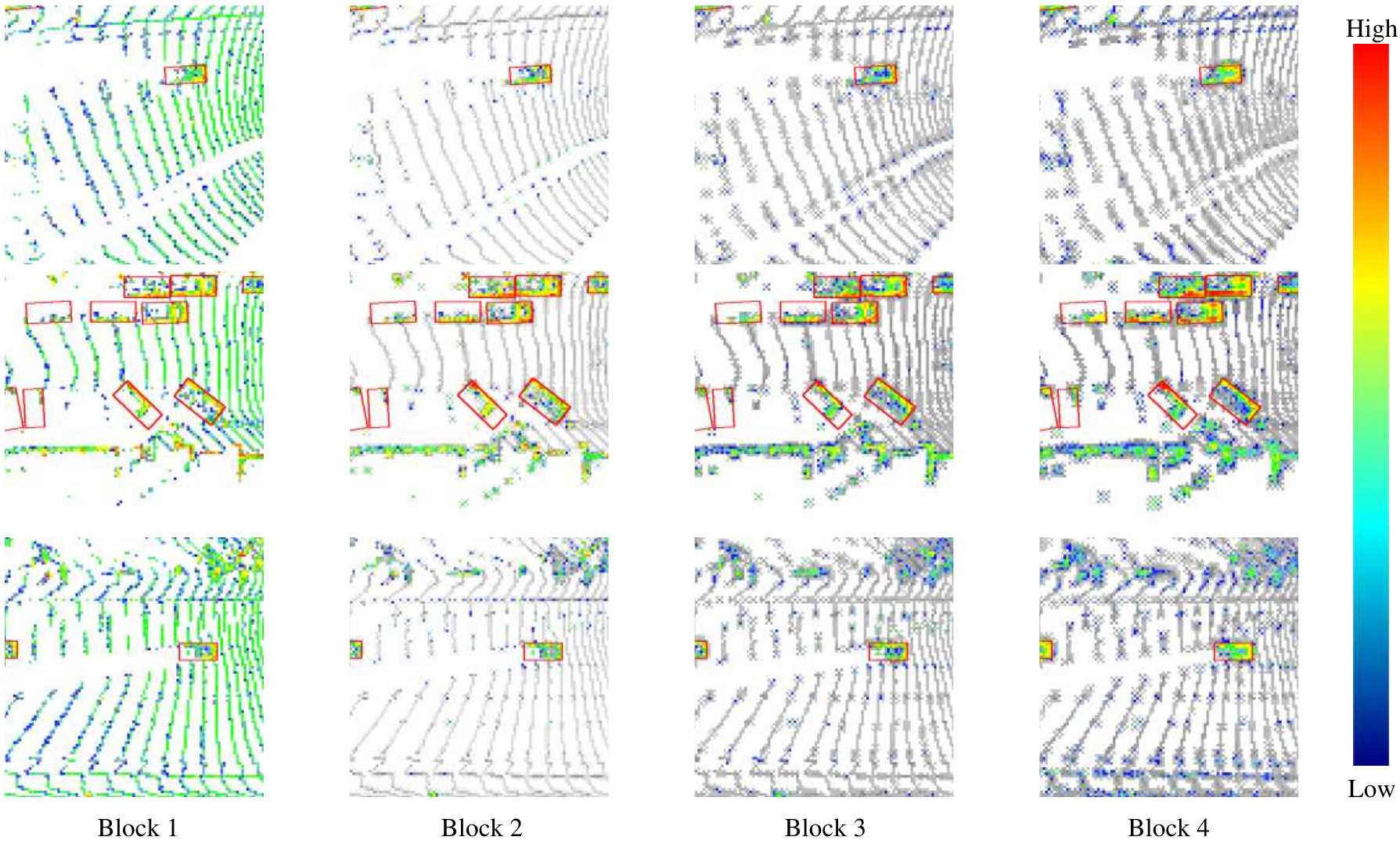}
    \caption{Visualization of feature map of different blocks. We highlight the foreground annotated by red GT boxes. The color map represents the magnitude of the feature response. }
    \label{fig_heatmap}
    \vspace{-10pt}
\end{figure}

\subsection{Visualization for Feature Map}
\label{sec:vis_feature}
As shown in Figure~\ref{fig_heatmap}, we visualize feature maps of different {\ourMethod} blocks. We can observe that as the features pass through more blocks, the magnitude of the foreground's feature response becomes larger, demonstrating the rationality of distinguishing foreground voxels by feature response. Besides, we find that the foreground features become more dense and more distinguished, which also demonstrates the effectiveness of the voxel generation operation.

\begin{figure}[t!]
    \centering
    \includegraphics[width=0.99\linewidth]{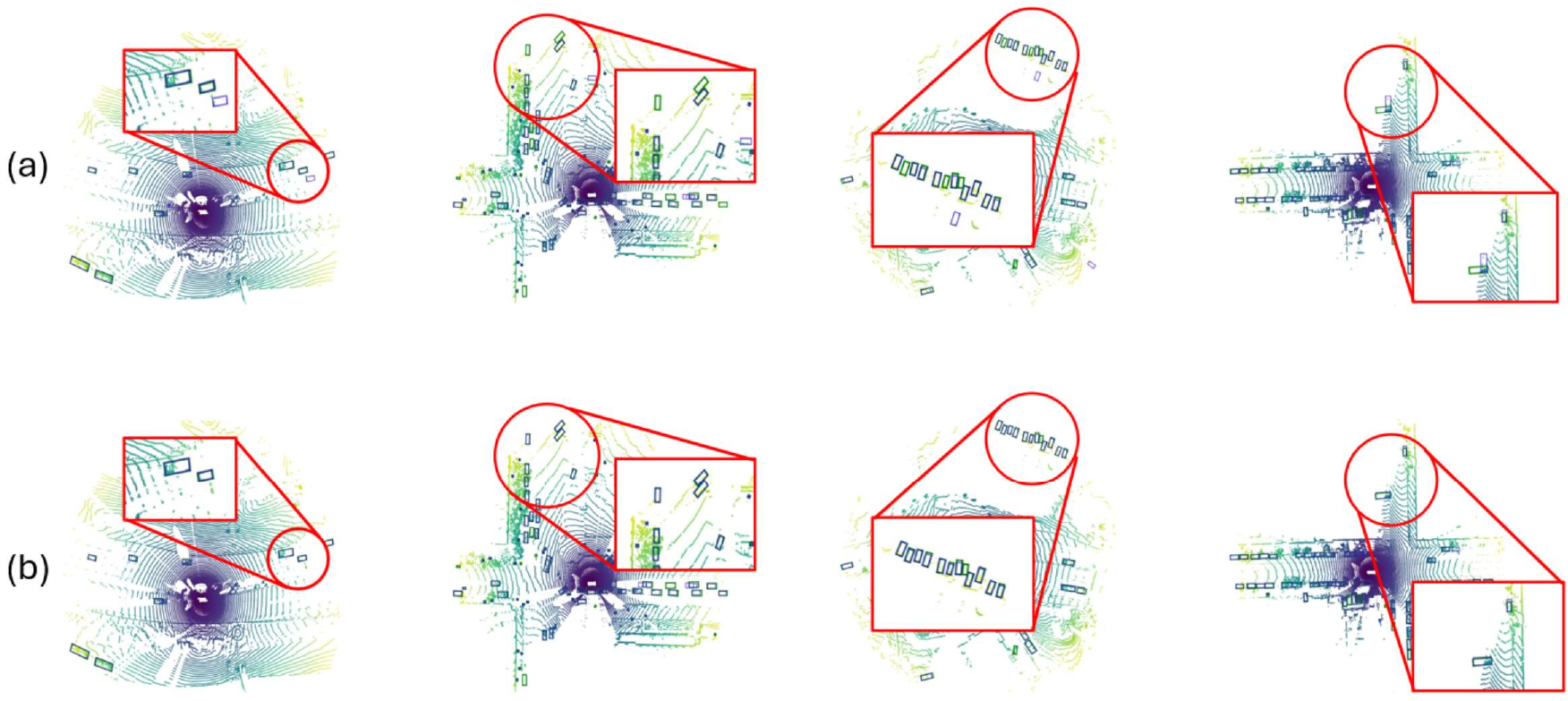}
    \caption{Comparison of DSVT and {\ourMethod} on the WOD \textit{validation} set from the BEV perspective. Blue and green boxes are the prediction and ground truth boxes. It can be seen that {\ourMethod} can achieve better results compared to DSVT, demonstrating the superiority of {\ourMethod}.}
    \label{fig_visualize}
    \vspace{-10pt}
\end{figure}

\subsection{Comparison of Qualitative Results with DSVT}
\label{sec:cmp_results}
To illustrate the superiority of {\ourMethod}, we present the visualization of the qualitative results of DSVT~\cite{wang2023dsvt}~(a) and {\ourMethod}~(b) on the WOD~\cite{sun2020scalability} \textit{validation} set, as shown in Figure~\ref{fig_visualize}. Specifically, in the first and third columns,  our {\ourMethod} can reduce more false positives compared with DSVT. In the second column, our {\ourMethod} even detects some hard objects at a distance. In the last column, our {\ourMethod} can achieve more accurate localization. These qualitative results demonstrate the superior performance of our {\ourMethod}.

\subsection{Qualitative Results}
\label{sec:qualitative}
\begin{figure}[htbp]
    \centering
    \includegraphics[width=0.99\linewidth]{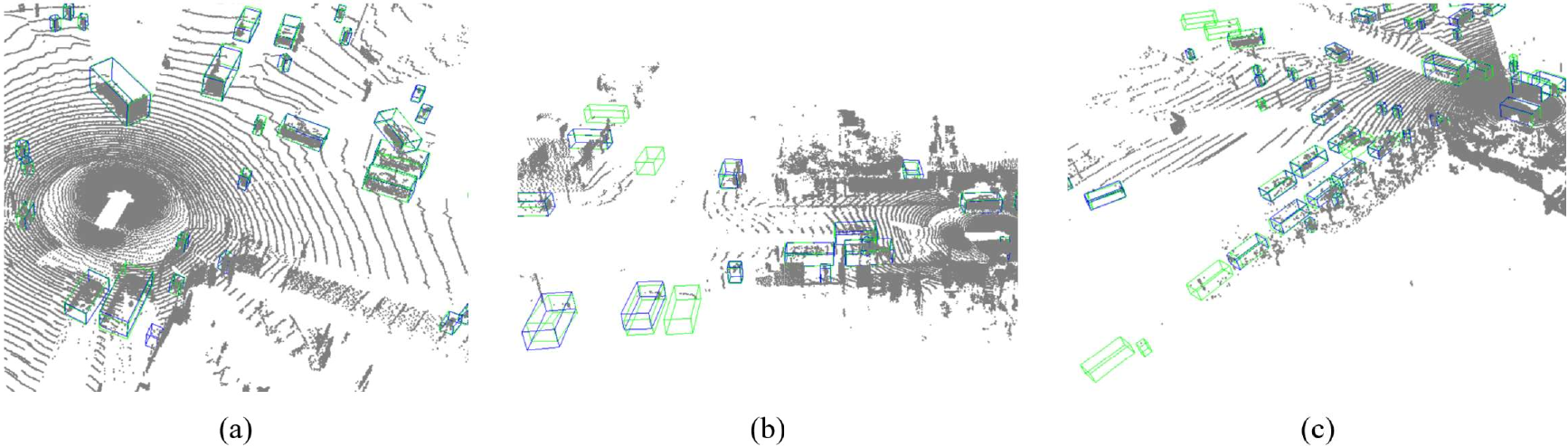}
    \caption{Qualitative results of {\ourMethod} on the WOD \textit{validation} set. Green and blue boxes denote ground truth and predicted bounding boxes, respectively. 
    }
    \label{fig_3d}
\end{figure}

As shown in Figure~\ref{fig_3d}, we visualize the qualitative results of {\ourMethod} on the WOD \textit{validation} set. As shown in the first column, {\ourMethod} can still achieve satisfactory results even in crowded 3D scenes. However, as shown in the second and third columns, {\ourMethod} misses some objects at a distance with sparse point clouds. Therefore, we will further improve the performance of distant objects by fusing the image features in the future. 

\subsection{Broader Impacts}
\label{sec:broader}
{\ourMethod} achieves promising performance for 3D object detection, enhancing the safety of autonomous driving. However,  {\ourMethod} has relatively high requirements on computing resources to achieve faster running speed, which puts forward higher requirements for the hardware of autonomous driving. Future research could focus on optimizing {\ourMethod} to improve bottlenecks in running speed while maintaining high detection accuracy, making it more accessible and practical for autonomous driving.


\clearpage

\end{document}